\documentclass{IEEEtran}
\usepackage{amsmath,amsfonts}
\usepackage{algorithmic}
\usepackage{algorithm}
\usepackage{booktabs}
\usepackage{array}
\usepackage[caption=false,font=normalsize,labelfont=sf,textfont=sf]{subfig}
\usepackage{textcomp}
\usepackage{stfloats}
\usepackage{url}
\usepackage{verbatim}
\usepackage{graphicx}
\usepackage{enumerate}
\usepackage{caption}
\usepackage{cite}
\usepackage{epstopdf}         
\usepackage{flushend}
\usepackage{hyperref}
\usepackage{doi}
\usepackage{multirow}
\usepackage[table,xcdraw]{xcolor} 
\usepackage{listings}
\usepackage{color}
\usepackage{float}
\definecolor{dkgreen}{rgb}{0,0.6,0}
\definecolor{gray}{rgb}{0.5,0.5,0.5}
\definecolor{mauve}{rgb}{0.58,0,0.82}

\begin{document}

\title{WTEFNet: Real-Time Low-Light Object Detection for Advanced Driver Assistance Systems}

\author{Hao Wu, Junzhou Chen, Ronghui Zhang, Nengchao Lyu, Hongyu Hu, Yanyong Guo, and Tony Z. Qiu

\thanks{Manuscript submitted May 26, 2025; (Corresponding author: Ronghui Zhang.)}

\thanks{This project is jointly supported by National Natural Science Foundation of China (Nos.52172350, 51775565), Guangdong Basic and Applied Research Foundation (No.2022B1515120072), Guangzhou Science and Technology Plan Project (No.2024B01W0079), Nansha Key RD Program (No.2022ZD014), Science and Technology Planning Project of Guangdong Province (No.2023B1212060029).}

\thanks{Hao Wu, Junzhou Chen and Ronghui Zhang are with the Guangdong Provincial Key Laboratory of Intelligent Transportation System, School of Intelligent Systems Engineering, Sun Yat-sen University, Guangzhou 510275, China (e-mail: wuhao225@mail2.sysu.edu.cn; chenjunzhou@mail.sysu.edu.cn; zhangrh25@mail.sysu.edu.cn).}
\thanks{Nengchao Lyu is with the Intelligent Transportation Systems Research Center, Wuhan University of Technology, Wuhan 430063, China. (e-mail: lnc@whut.edu.cn).}
\thanks{Hongyu Hu is with State Key Laboratory of Automotive Simulation and Control, Jilin University, Changchun 130022, China (e-mail: huhongyu@jlu.edu.cn)}
\thanks{Yanyong Guo is with the School of Transportation, Southeast University, Nanjing 210097, China. (guoyanyong@seu.edu.cn).}
\thanks{Tony Z. Qiu is with Department of Civil and Environmental Engineering, University of Alberta, Edmonton, Alberta, Canada(e-mail: zhijunqiu@ualberta.ca).}

}
%

\maketitle
\begin{abstract}
Object detection is a cornerstone of environmental perception in advanced driver assistance systems(ADAS). However, most existing methods rely on RGB cameras, which suffer from significant performance degradation under low-light conditions due to poor image quality. To address this challenge, we proposes \textbf{WTEFNet}, a real-time object detection framework specifically designed for low-light scenarios, with strong adaptability to mainstream detectors. WTEFNet comprises three core modules: a \textbf{Low-Light Enhancement (LLE)} module, a \textbf{Wavelet-based Feature Extraction (WFE)} module, and an \textbf{Adaptive Fusion Detection (AFFD)} module. The LLE enhances dark regions while suppressing overexposed areas; the WFE applies multi-level discrete wavelet transforms to isolate high- and low-frequency components, enabling effective denoising and structural feature retention; the AFFD fuses semantic and illumination features for robust detection. To support training and evaluation, we introduce \textbf{GSN}, a manually annotated dataset covering both clear and rainy night-time scenes. Extensive experiments on BDD100K, SHIFT, nuScenes, and GSN demonstrate that WTEFNet achieves state-of-the-art accuracy under low-light conditions. Furthermore, deployment on a embedded platform (NVIDIA Jetson AGX Orin) confirms the framework's suitability for real-time ADAS applications.
\end{abstract}


\section{Introduction}
\IEEEPARstart{A}{s} a core element in autonomous driving systems, object detection is responsible for predicting the positions and sizes of surrounding vehicles\cite{cai2021yolov4,wang2024yolov8,ye2023real,zhang2023msffa}, pedestrians\cite{yuan2023triangular,hsu2022pedestrian}, traffic lights\cite{zhao2024miaf,tabernik2019deep}, and obstacles\cite{darms2009obstacle}. This technology provides accurate environmental information crucial for intelligent route planning and collision avoidance. It plays an essential role in the perception module of autonomous driving, supporting vital driving decisions and ensuring higher vehicle safety within Advanced Driver Assistance Systems (ADAS).

Under conditions of limited or no illumination, both visibility and object recognition performance are seriously degraded, greatly compromising nighttime driving safety. According to studies by the International Commission on Illumination (CIE), nighttime traffic accident rates are three times higher than during the day, with more severe consequences and greater threats to life and property. Therefore, ADAS must integrate more advanced onboard embedded systems to improve low-light object detection capabilities.

As illustrated in Figure 1, low-light object detection systems typically consist of three main components: object detection, edge computing, and decision making. Initially, the object detection module utilizes sensors such as high-resolution cameras to continuously acquire real-time information regarding the position and dimensions of surrounding objects. Next, the edge computing module utilizes either onboard devices or cloud-based resources to further process and analyze the detection results, providing a comprehensive assessment of both the immediate traffic environment and the vehicle’s operational status. Finally, the decision making module uses these assessments to determine optimal driving strategies, including steering, braking, and acceleration, thereby maximizing vehicle safety.

Numerous studies have focused on image-based\cite{shirmohammadi2014camera} object detection, which is commonly categorized into conventional algorithms and deep learning-based algorithms. Traditional methods typically rely on manually crafted features for object detection, such as HOG (Histogram of Oriented Gradients)\cite{wang1999image,shen2011adaptive}. However, these approaches are limited by environmental factors like imaging conditions, object deformation and occlusion, resulting in weaker feature representation and limited generalization. Recent developments in deep learning have led to significant progress in object detection algorithms, greatly enhancing both accuracy and inference speed. These methods offer unparalleled advantages over traditional approaches, making them the dominant choice for object detection.

\begin{figure*}
    \centering
    \includegraphics[scale=0.93]{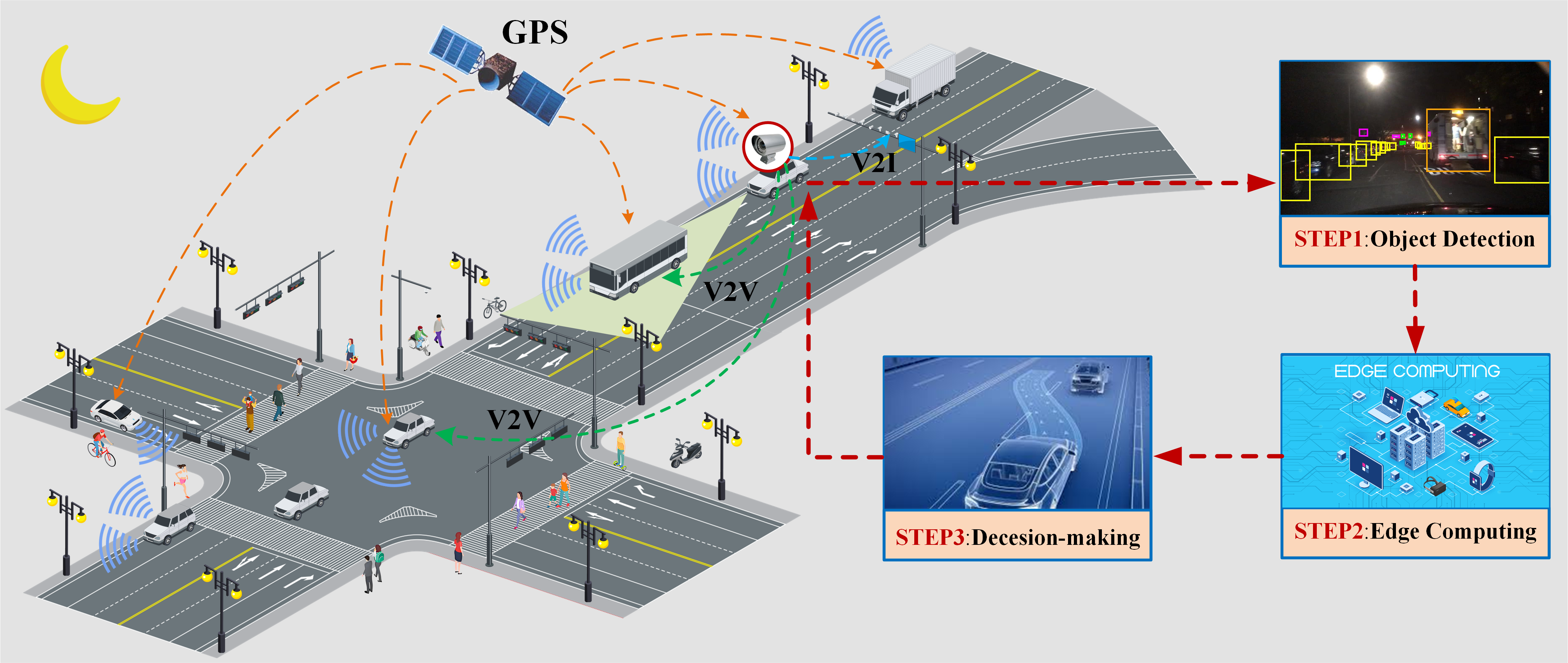}
    \caption{Low-light object detection for ADAS\cite{uniadslides,csdn2022,baiduwenku2025,googleimages2025}.}
    \label{fig:vechicle}
\end{figure*}

Deep learning-based object detection approaches are typically classified into two-stage methods, which utilize Region Proposal Networks (RPN), and one-stage methods, which rely on regression-based methods. Representative two-stage methods include R-CNN\cite{girshick2014rich}, SPPNet, Fast R-CNN\cite{girshick2015fast}, and Faster R-CNN\cite{ren2015faster}, while widely used one-stage models include the YOLO\cite{redmon2016you} series, SSD\cite{liu2016ssd,ni2023improved} series, and RetinaNet. Although these algorithms have achieved notable performance in object detection, certain challenges persist.

Under ideal road conditions, state-of-the-art object detection methods usually yield satisfactory results. However, in low-light or dim environments, objects surrounding the vehicle may appear blurred due to underexposure, making it difficult to extract effective features. Additionally, in urban nighttime scenarios, the glare from high-power vehicle headlights and streetlights poses even greater challenges for multi-object detection tasks. Therefore, improving object detection performance under low-light conditions is an urgent need.

The majority of studies utilize low-light enhancement algorithms as a preprocessing module, followed by detection algorithms to locate objects within the enhanced images. Traditional low-light image enhancement methods mainly include Retinex theory, histogram equalization, and HDR algorithms. Although these methods can achieve a certain degree of enhancement, they often suffer from issues such as color distortion and noise amplification, making them less adaptable to the wide variety of low-light natural scenes and limiting their practical applications. 

In contrast, deep learning-based low-light enhancement methods demonstrate superior performance and greater adaptability across different scenes. MSRNet\cite{jobson1997multiscale} was the first algorithm to combine Retinex theory with convolutional neural networks (CNNs), replacing the Gaussian convolution kernel in the MSR enhancement algorithm with a CNN to directly learn an end-to-end mapping between low-light and bright images. To improve the model's noise suppression capability, some studies have added various types of noise reduction modules. KinD\cite{zhang2019kindling} uses a Restoration-Net to perform denoising on the reflectance image, while LLNet is designed to adapt to diverse lighting and noise conditions.

Although these algorithms have achieved considerable enhancement effects, creating paired datasets is costly, and existing paired datasets only cover a limited range of scenes. Most low-light images captured in natural environments lack corresponding reference images. To improve the generalization ability of the algorithms, some research has adopted unsupervised methods to train models on such datasets. RetinexDIP\cite{zhao2021retinexdip}, for example, requires only a single low-light image for iterative training, addressing the challenges of data collection. The zero-shot learning model, Zero-DCE\cite{li2021learning}, further reduces dependence on paired or unpaired datasets, though its enhanced images still contain noticeable grayish-white noise. From the studies above, it is clear that the primary challenges for current low-light image enhancement methods remain the difficulty of collecting paired datasets and the introduction of noise.

The preceding sections have thoroughly examined the key challenges associated with object detection in low-light environments. In general, there is an urgent need for an innovative approach that not only suppresses noise and enhances image quality under poor illumination but also integrates effectively with mainstream object detection models. To address the issues outlined above, we propose a real-time object detection framework based on wavelet transform, called WTEFNet. As a novel framework, WTEFNet is designed to improve object detection performance under low-light conditions. The following sections will detail the distinct contributions of this work, highlighting its potential to drive meaningful progress in ADAS development under low-light conditions. The primary contributions of this study are summarized as follows:

\begin{enumerate}[(1)]
\item {
We present WTEFNet, a real-time object detection framework designed for low-light environments. It exhibits high adaptability to mainstream detection networks and significantly improves detection accuracy under challenging illumination conditions.
}

\item {
We develop a wavelet-based feature extraction (WFE) module and an adaptive fusion detection (AFFD) module. The WFE separates high- and low-frequency components to reduce noise and retain salient features, while the AFFD adaptively fuses semantic and illumination information to enhance detection robustness.
}

\item {
To address the lack of annotated datasets for low-light object detection, we construct GSN, a dedicated benchmark comprising 2,031 manually labelled images captured in urban night-time scenes across Guangzhou and Shanghai, including both clear and rainy conditions to ensure diversity.
}

\item {
Extensive experiments on BDD100K, SHIFT, nuScenes, and GSN demonstrate that WTEFNet achieves superior performance compared to existing methods in low-light scenarios. Moreover, it supports real-time inference on resource-constrained embedded platforms such as the NVIDIA Jetson AGX Orin.
}
\end{enumerate}

The remainder of this paper is structured as follows: Section \uppercase\expandafter{\romannumeral2} provides a review of related literature. Section \uppercase\expandafter{\romannumeral3} introduces the architecture of the proposed WTEFNet for low-light object detection. Section \uppercase\expandafter{\romannumeral4} presents the experimental evaluation, and Section \uppercase\expandafter{\romannumeral5} summarizes the conclusions of this work.

\section{Related Work}
\subsection{Low-light Image Enhancement}For low-light image enhancement tasks, early studies primarily relied on traditional methods, such as histogram equalization-based techniques. Representative examples include AHE\cite{pizer1987adaptive}, DSIHE\cite{wang1999image}, and IBBHE\cite{shen2011adaptive}. These methods are simple in design and computationally efficient, and they can produce acceptable enhancement results under certain conditions. However, they often suffer from problems such as color distortion and noise amplification in the enhanced images.

In addition, Retinex theory has been widely applied to low-light image enhancement. The SSR\cite{jobson1997properties} algorithm was the first to propose using a single-scale Gaussian surround function to estimate the illumination component from the original image, thereby achieving enhancement. BIMEF\cite{ying2017bio} addressed issues such as overly strong contrast and over-enhanced brightness by incorporating concepts from camera response curves and human visual systems. Building on BIMEF, CRM\cite{chang2013improved} estimated camera response functions from multiple exposure image sets to further alleviate color and brightness distortions.

With the rapid development of hardware computing capabilities, a growing number of studies have focused on deep learning-based methods for low-light image enhancement. Depending on the training strategy, these methods can be broadly categorized into paired and unpaired training approaches.

\subsubsection{trained with paired data} 

MSRNet replaced the Gaussian convolution kernel in the MSR\cite{shen2017msr} algorithm with a convolutional neural network, enabling end-to-end learning from low-light to normal-light images. However, the resulting model lacks noise suppression capabilities. LightenNet\cite{li2018lightennet}, based on Retinex theory, fused illumination and reflectance for enhancement, but still struggled to effectively suppress noise. To address this issue, various models integrated noise suppression modules, such as RetinexNet\cite{wei2018deep}, KinD\cite{zhang2019kindling}, and KinD++\cite{zhang2021beyond}. LLNet\cite{lore2017llnet} was one of the first to apply stacked sparse denoising autoencoders (SSDA) to natural low-light image enhancement, providing both contrast enhancement and denoising while maintaining good generalizability.

\subsubsection{trained with unpaired data}

Creating paired datasets is often costly, and existing paired datasets typically cover only a limited range of scenes. In real-world environments, most low-light images captured in natural settings lack corresponding reference images. As a result, many methods adopt unsupervised training strategies. For example, SID-NISM\cite{zhang2020sid} performs self-supervised training using only low-light images, while RetinexDIP\cite{zhao2021retinexdip} iteratively trains on a single low-light image, effectively addressing the data collection challenge. Other approaches \cite{ignatov2018wespe,jiang2021enlightengan} design both global and local adversarial generative networks (GANs)\cite{goodfellow2014generative} to perform adaptive illumination correction for low-light images. SRANet\cite{yang2021lowlight}, built upon EnlightenGAN\cite{jiang2021enlightengan}, introduces an optimized adversarial denoising mechanism, improving the model’s ability to suppress noise in real low-light images. DRBN\cite{yang2021band} consists of two stages: recursive band learning and band recombination. It is compatible with both paired and unpaired datasets and performs competitively in contrast enhancement and structural detail recovery. Zero-DCE\cite{li2021learning} adopts a zero-shot learning strategy that relies solely on parameter maps estimating pixel-wise mapping curves for iterative enhancement, achieving high runtime efficiency. SCI\cite{ma2022toward} is another lightweight model that enhances illumination efficiently by employing a multi-stage self-calibrated illumination module. UIA\cite{wang2024unsupervised} introduces two specialized strategies to improve illumination restoration and feature alignment, significantly enhancing the performance of downstream tasks in a plug-and-play manner. This method further reduces reliance on paired or unpaired datasets for image enhancement. 

Given the success of Transformers\cite{vaswani2017attention} in various vision applications, several recent methods have explored their use in low-light image enhancement, such as Retinexformer\cite{cai2023retinexformer} and UPT-Flow\cite{xu2025upt}. Although these models achieve impressive enhancement results, they tend to have high computational complexity, limiting their applicability in time-sensitive downstream tasks. IAT\cite{cui2022you} addresses this limitation by proposing a lightweight and real-time enhancement network, which uses an end-to-end Transformer architecture to mitigate the adverse effects of poor lighting on visual tasks.

FFENet\cite{liu2024ffenet} is a novel low-light image enhancement method that improves enhancement accuracy by separately processing high-frequency and low-frequency components based on their information characteristics. To address the limitations of pixel-level reconstruction loss in learning accurate mappings from low-light to normal-light images, GACA\cite{yao2024gaca} introduces a novel regularization mechanism to refine this mapping process. Extensive experiments on benchmark datasets and downstream segmentation tasks demonstrate that GACA achieves excellent performance.

\subsection{Object Detection} Current mainstream object detection methods can generally be broadly divided into two classes: two-stage detectors and one-stage detectors.
\subsubsection{Two-Stage Detectors}
Two-stage detectors typically first generate a set of region proposals, which are then classified and refined during the subsequent stage. R-CNN\cite{girshick2014rich} introduces this approach by using selective search for proposal generation and CNNs\cite{krizhevsky2012imagenet} for feature extraction, but suffers from high computational cost due to repeated feature computation. SPPNet\cite{he2015spatial} improves efficiency via spatial pyramid pooling, enabling shared feature maps, but lacks end-to-end training. Fast R-CNN\cite{girshick2015fast} addresses these issues with RoI pooling and a unified training process, greatly improving speed and accuracy. Faster R-CNN\cite{ren2015faster} further enhances performance by integrating a learnable Region Proposal Network (RPN), achieving end-to-end optimization and becoming one of the most widely adopted detectors.

\subsubsection{one-Stage Detectors}
One-stage detectors eliminate the proposal generation step and perform detection in a single pass, enabling faster inference. Among them, SSD\cite{liu2016ssd} makes a major contribution by detecting objects of various sizes through multiple feature maps at different scales. While SSD is fast and easy to train, it performs poorly on very small or densely packed objects and is sensitive to anchor box configurations. RetinaNet\cite{lin2017focal} builds a one-stage architecture on top of a Feature Pyramid Network and introduces Focal Loss to address the foreground–background class imbalance problem.
CenterNet\cite{law2018cornernet} and CornerNet\cite{zhou2019objects} formulate object detection as a keypoint regression problem. These models are anchor-free and do not require NMS, resulting in a relatively simple structure. FCOS\cite{tian2019fcos} also avoids the use of anchors, reducing dependence on manually designed hyperparameters and offering better scalability. The YOLO\cite{redmon2016you,wang2024yolov10,tian2025yolov12} series has been one of the most widely adopted and continuously improved families of object detectors. With successive versions, it offers strong real-time performance, a simple architecture, and is highly deployment-friendly. In recent years, Transformers have significantly influenced the field of computer vision. DETR\cite{carion2020end} reframes object detection as a set prediction problem by introducing a transformer-based encoder–decoder architecture. It enables end-to-end object detection without relying on NMS. However, DETR suffers from slow convergence and high computational cost. Subsequent methods such as Deformable DETR\cite{zhu2020deformable} and DINO\cite{zhang2022dino} aim to alleviate these limitations. Although these approaches have improved efficiency and detection accuracy, they still face challenges in satisfying the strict requirements of real-time detection scenarios.

\begin{figure*}[t]
    \centering
    \includegraphics[scale=0.85]{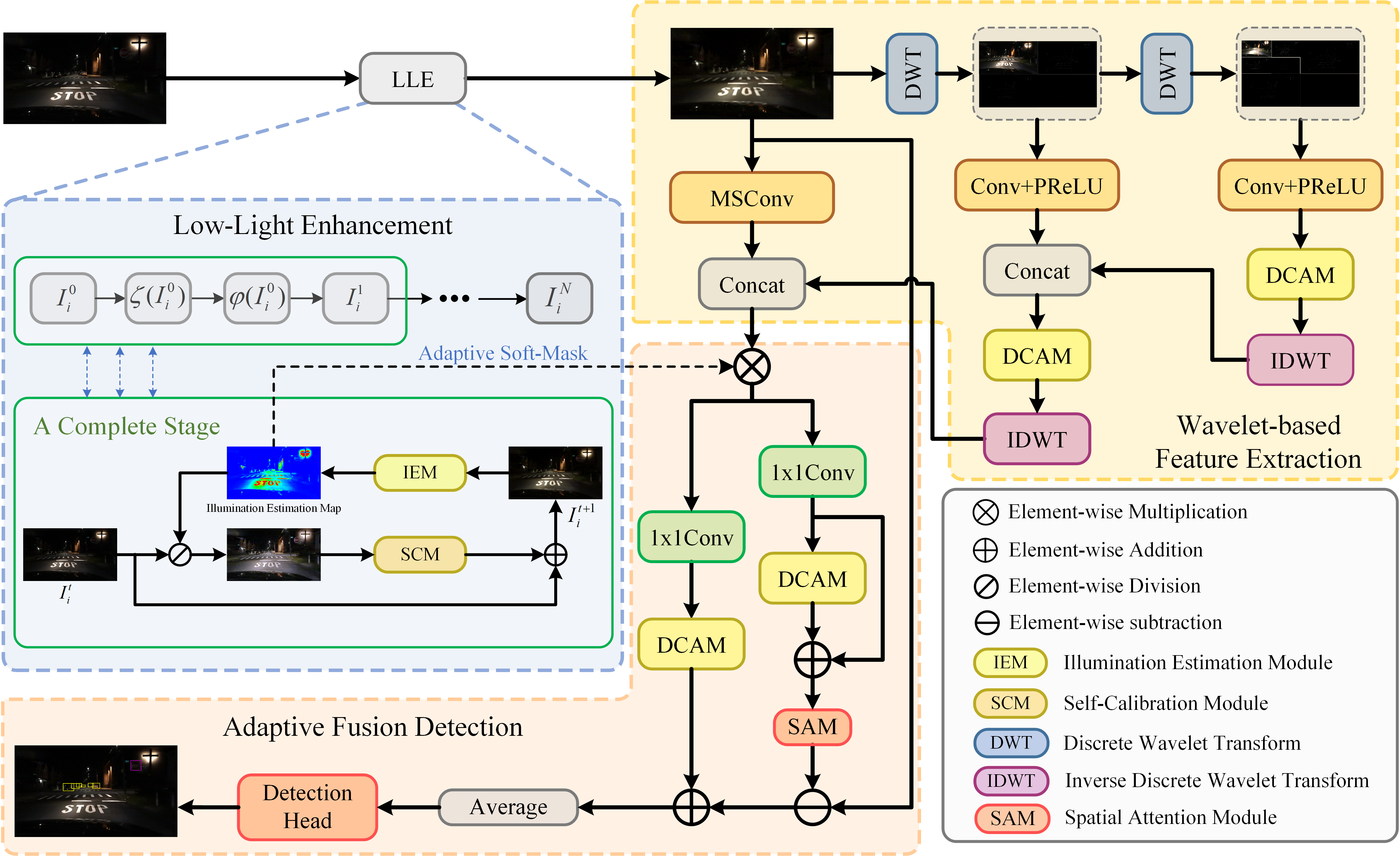}
    \caption{The overall architecture of WTEFNet is illustrated in the schematic diagram, which consists of three main components: a low-light enhancement module, a wavelet-based feature extraction module, and an adaptive fusion detection module. The full names of all common symbols and structures are provided in the legend at the bottom-right corner. The detailed structures of DCAM and MsConv are shown in Figure \ref{fig:DCAM} and Figure \ref{fig:MSConv}, respectively.}
    \label{fig:WDLEDet}
\end{figure*}

\subsection{Object Detection under Low-Light Conditions}

Low-light object detection methods can be broadly categorized into three groups: enhancement-then-detection, enhancement-for-detection, and detector learning strategies under low-light conditions. The first category focuses on improving image brightness using low-light enhancement techniques, thereby indirectly increasing detection accuracy. The second category focuses on jointly optimizing the enhancement and detection modules through unified end-to-end training, with the goal of simultaneously improving image quality and enhancing the effectiveness of feature representations for object detection. For example, PIA\cite{ma2022pia} proposes a parallel architecture that simultaneously addresses the requirements of both enhancing low-light images and performing object detection. \cite{sun2022rethinking} proposed an adversarial attack strategy inspired by the ADAM optimizer, which not only generates high-quality restored images but also significantly improves object detection accuracy. The third category adopts weakly supervised or unsupervised learning strategies to improve low-light object detection performance. Representative approaches include DA F-RCNN\cite{chen2018domain}, TDD\cite{he2022cross}, AT\cite{li2022cross}, UMT\cite{deng2021unbiased}, DAI-Net\cite{du2024boosting}, 2PCNet\cite{kennerley20232pcnet}, and WSA-YOLO\cite{hui2024wsa}, which focus on domain adaptation or self-supervision to bridge the gap between low-light and normal-light domains. The proposed low-light detection framework in this paper combines the advantages of the first two categories: It enhances the image while simultaneously extracting effective features to achieve better detection accuracy.

\section{Method}
\subsection{Overall Architecture}

\begin{figure}
    \centering
    \includegraphics[scale=0.75]{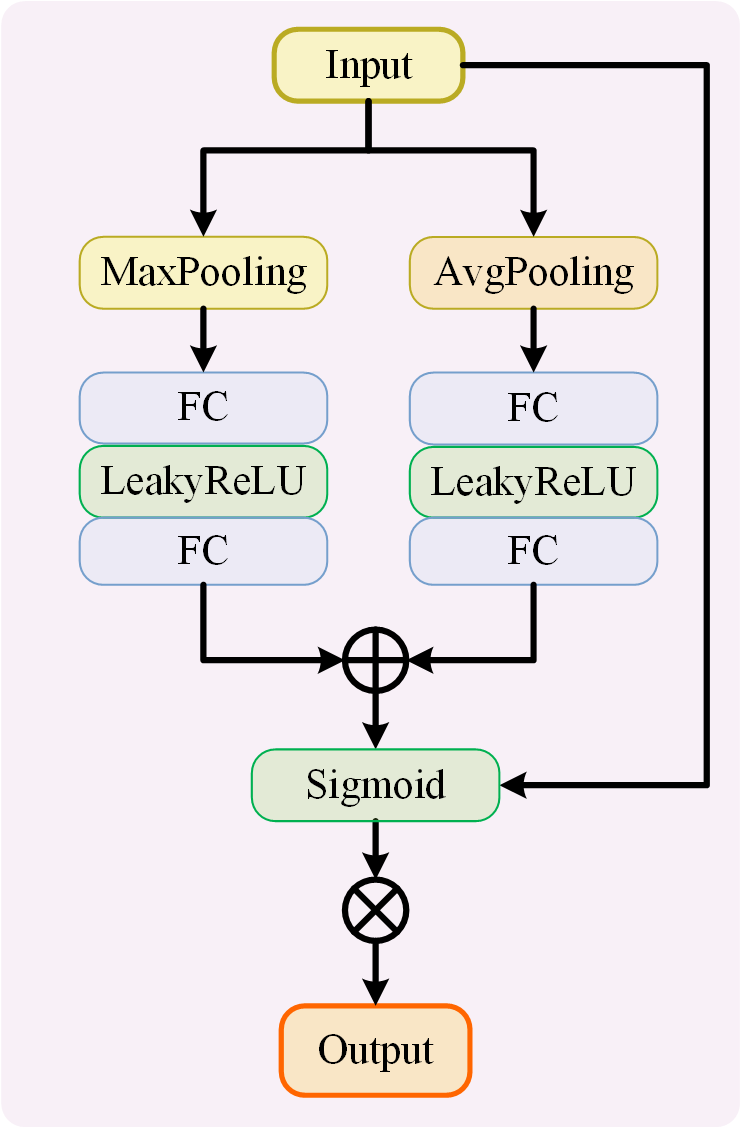}
    \caption{Densely Channel Attention Module(DCAM). }
    \label{fig:DCAM}
\end{figure}

\begin{figure}
    \centering
    \includegraphics[scale=0.70]{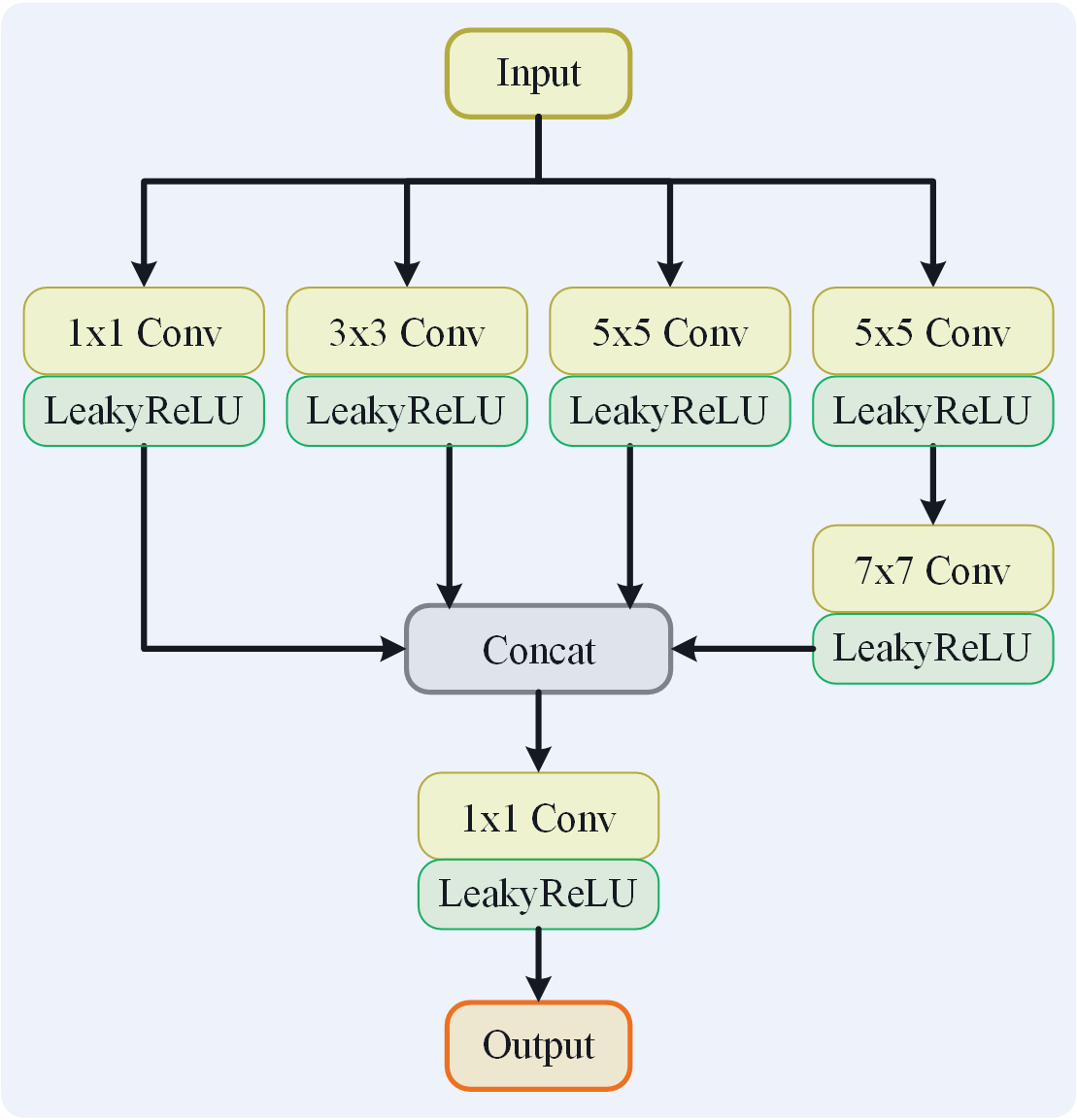}
    \caption{Multi-scale Convolution(MSConv) Block.}
    \label{fig:MSConv}
\end{figure}

In this work, we introduce a novel object detection framework tailored for low-light environments, named WTEFNet, as shown in Fig. \ref{fig:WDLEDet}, which comprises three primary modules: a low-light image enhancement module, a wavelet-based feature extraction module, and an adaptive fusion detection module. The full names of all common symbols and structures in Fig. \ref{fig:WDLEDet} are provided in the legend at the bottom right. The detailed designs of DCAM and MsConv are shown in Fig. \ref{fig:DCAM} and Fig. \ref{fig:MSConv}, respectively. The low-light image enhancement module integrates a lightweight enhancement method optimized for object detection. It restores distinctive visual features often lost under poor lighting by enhancing illumination in dark regions and self-calibrating overexposed areas. The wavelet-based feature extraction module, inspired by the UNet architecture, employs a multi-level discrete wavelet transform (DWT) to decompose images into high- and low-frequency components. After each transformation level, the feature resolutions are reduced to $\tfrac{H}{2} \times \tfrac{W}{2}$ and $\tfrac{H}{4} \times \tfrac{W}{4}$, respectively. These components are processed separately using DCAM and other techniques. The discrete inverse wavelet transform(IDWT) is then used to restore the feature maps to their original resolution. This design is intended to reduce noise and preserve salient object features. Finally, the adaptive fusion detection module integrates the features from the previous two modules to enable real-time and accurate object detection. It is compatible with various mainstream detectors, including the YOLO series. The detailed structures of each module will be elaborated in subsequent sections.

\subsection{Low-Light Enhancement}

According to the preceding analysis, low-light image enhancement can improve image quality and thereby boost the accuracy of object detection. However, under actual road conditions—particularly in urban environments—onboard cameras are also prone to glare at night, caused by oncoming vehicle headlights or high-intensity street lights. In such scenarios, In addition to enhancing underexposed regions, it is essential to suppress overexposure in bright areas. To address this, we design a stage-wise low-light enhancement module capable of handling both challenges.

Based on Retinex theory, the relationship between a low-light image $ I $ and its enhanced image $I'$ is expressed as $ I = I' \otimes x $, Consequently, the enhanced image can be derived by $ I' = I \oslash x $, where $x$ denotes the illumination map, which is typically regarded as the core component to be optimized in low-light enhancement tasks. Inaccurate estimation of illumination may lead to either insufficient or excessive enhancement. To balance accuracy and efficiency, we design a stage-wise illumination estimation strategy inspired by SCI \cite{ma2022toward}, as shown in the upper left part of Figure \ref{fig:WDLEDet}. This approach integrates the self-calibration module at each stage to enable more robust and efficient illumination refinement. The enhancement module is mathematically formulated as follows
\begin{equation}
    I_i^{n+1} = \mathcal{C}(\mathcal{F}(I_i^n)), 
\end{equation}
where $n$ denotes the stage($n = 0,...,N$), $\mathcal{C}$ and $\mathcal{F}$ represent self-calibration module and illumination estimation module, respectively. In stage n, $\mathcal{F}$ is formulated by
\begin{equation}
    \mathcal{F}(I_i^n) = I_i^n \oslash x_i^n, 
\end{equation}
where illumination component $x_i^n$ is estimated by the trainable networks. $\mathcal{C}$ is formulated by
\begin{equation}
    \mathcal{C}(\mathcal{F}(I_i^n)) = 
\begin{cases} 
R_i^n = \mathcal{G}(\mathcal{F}(I_i^n)), \\
I_i^{n+1} = I_i^n \oplus R_i^n,
\end{cases}    
\end{equation}
where $\mathcal{G}$ represents the network responsible for generating calibrated residual map $R_i^n$. In practice, the calibration module at each stage progressively refines the input of that stage, thereby indirectly influencing its output.  

Although low-light image enhancement modules can effectively improve image quality, some underexposed and overexposed regions often remain. These regions disrupt the continuity of local features, thereby impairing the model’s ability to accurately detect objects. During the enhancement process, we compute an illumination map $x$, which determines the light enhancement ratio for each pixel in the original image. Empirically, underexposed regions require a higher enhancement ratio, while overexposed regions require a lower one. This ratio can be used to estimate the relative importance of each image region during training. To address this, we propose an Adaptive Soft-Mask mechanism, inspired by the STEPS\cite{zheng2023steps}. Specifically, we generate a confidence map $M_c$ from the illumination map $x$, assigning lower confidence values to both underexposed and overexposed regions. The confidence map $M_c$ is defined as follows
\begin{equation}
    M_c = 
\begin{cases}
\frac{1}{\sqrt{1+c_1^2(x-l)^2}}, & x_{min}\leq x\leq l \\
1,                      & l\leq x\leq r \\
\frac{1}{\sqrt{1+c_2^2(x-r)^2}}, & r\leq x\leq x_{max}
\end{cases}
\end{equation}
where $c_1$ and $c_2$ are the coefficients, $l$ and $r$ denote the lower and upper bounds, respectively. These values are obtained through statistical computation over the entire dataset.

\subsection{Wavelet-based Feature Extraction}
To effectively suppress noise introduced by both the input image itself and the low-light enhancement module, we design a Wavelet-based Feature Extraction(WFE) module. This module adopts a U-shaped self-guided architecture to better leverage the multi-scale information of the image. Information extracted at lower resolutions is progressively propagated to higher-resolution sub-networks, serving as guidance for the feature extraction process. WFE applies Discrete Wavelet Transform (DWT) to the enhanced image to generate multi-scale feature maps. These low-resolution feature maps are first processed through basic convolutional layers and activation functions, and then passed through a Densely Channel Attention Module (DCAM). The resulting features are restored to a higher resolution using Inverse Discrete Wavelet Transform (IDWT) and subsequently fused with the features from the preceding scale.At the highest resolution level, the enhanced image is fed into a Multi-scale Convolution (MSConv) Block to obtain the initial features, which are then concatenated with the wavelet-processed features to generate denoised feature representations. The overall process can be formally expressed as
\begin{equation}
F_o =  DWT(I_e) = [F_1,F_2,F_3,F_4], F_o\in R^{\tfrac{H}{2} \times \tfrac{W}{2} \times 4}
\end{equation}
\begin{equation}
F'_o=PReLU(Conv(F_o)),
\end{equation}
\begin{equation}
F_s = [F_{11},F_{12},F_{13},F_{14}] = DWT(F_1),F_s \in R^{\tfrac{H}{4} \times \tfrac{W}{4}\times 4}
\end{equation}
\begin{equation}
F'_s=DCAM(PReLU(Conv(F_s))),
\end{equation}
\begin{equation}
F''_o=concat(F'_o,IDWT(F'_s)),
\end{equation}
\begin{equation}
F_d=concat(MsConv(I_e),IDWT(DCAM(F''_o))),
\end{equation}
where $I_e$ denotes the enhanced image. $F_o$ and $F_s$ represent the features obtained from the first and second levels of discrete wavelet transform (DWT), respectively.$F_d$ represents the final fused feature.

\textbf{Densely Channel Attention Module(DCAM)}\quad As illustrated in Fig. 3, the DCAM adopts a dual-branch architecture, where one branch applies global average pooling, while the other uses max-average pooling. After the pooling operations, the features are passed through two fully connected layers and an activation function. To reduce computational cost, both branches share the same weights for their fully connected layers.

\textbf{Multi-scale Convolution(MSConv) Block}\quad Inspired by the Inception\cite{szegedy2017inception} architecture, the MsConv block consists of four parallel branches, each employing convolutional layers with different receptive fields to extract multi-scale features. The last branch incorporates two consecutive convolution layers followed by an activation function. Once multi-scale features are extracted, the outputs from each branch are concatenated and subsequently processed by a $1\times1$ convolution layer for feature dimension reduction.

\subsection{Adaptive Feature Fusion Detection}
The Adaptive Feature Fusion and Detection (AFFD) module further processes the previously enhanced and denoised features and outputs the final detection results. Specifically, the feature map $F_d$ is first element-wise multiplied with the confidence map $M_c$ and the resulting map is concatenated with $F_d$ to obtain $F'_d$. This feature is then fed into a dual-branch structure.

The first branch consists of a $1 \times 1$ convolution layer followed by the DCAM. The second branch builds upon the first by incorporating residual connections and adding a spatial attention module. Notably, although the two branches share a similar structure, their parameters are not shared. The output of the second branch is subtracted from the initially enhanced image $I_e$ to compute a residual, which is then averaged with the output of the first branch to produce the final feature map $F_f$. This feature map is subsequently fed into the detection head to generate the detection results. The choice of detection head is flexible; we recommend using detectors from the YOLO series. In our experiments, we adopt YOLOv10 and YOLOv12 as representative detection heads.

\subsection{Loss Function}
Overall, the loss function for the low-light object detection task consists of two components, which can be computed as
\begin{equation}
\mathcal{L}_{total} = \alpha\mathcal{L}_{det}\otimes{M_c}  + \beta\mathcal{L}_{lle},
\end{equation}
where $\mathcal{L}_{det}$ and $\mathcal{L}_{lle}$ represent the detection loss and the image enhancement loss, respectively. The definition of $\mathcal{L}_{det}$ depends on the selected detection head and is therefore not elaborated further in this section. 

The enhancement loss $\mathcal{L}_{lle}$ can be defined as
\begin{equation}
\mathcal{L}_{lle} = \zeta\mathcal{L}_{f} + \eta \mathcal{L}_{s},
\end{equation}
where $\mathcal{L}_{s}$ and $\mathcal{L}_{f}$ represent the smoothing and fidelity loss, respectively. They are formulated as
\begin{equation}
\mathcal{L}_f=\frac{1}{|I|}\sum_{I}\sum_n||x^n-I^n||_2
\end{equation}

\begin{equation}
\mathcal{L}_{s}=\frac{1}{|I|}\sum_{I}\sum_n\sum_i\sum_{j\in\mathcal{N}(i)}\omega_{i,j}|x^n(i)-x^n(j)|
\end{equation}
where $n$ is the stage, and $i$ denotes the $i$-th pixel in the image. $\mathcal{N}(i)$ represent the adjacent pixels in the $5 \times 5$ window centered at pixel $i$. $\omega_{i,j}$ denotes the weight coefficient of the Gaussian kernel, which is formulated as 

\begin{equation}
w_{i,j}  =\exp\left(-\frac{\sum_c(({I}_{i,c}+{R}_{i,c}^{n-1})-({I}_{j,c}+{R}_{j,c}^{n-1}))^2}{2\sigma^2}\right)
\end{equation}
where $c$ represents the number of image channels, and $\sigma$ represents the standard deviation of the Gaussian kernel function. In fact, $\mathcal{L}_{s}$ serves as a consistency regularization loss.

\section{Experiment}
\subsection{Datasets}
\textbf{BDD100K}
The BDD100K\cite{yu2020bdd100k} dataset is a comprehensive benchmark for autonomous driving research, comprising 100,000 video clips recorded under diverse driving conditions, with approximately 20\% captured at night. It provides extensive annotations for object detection tasks, covering 10 categories such as cars, buses, trucks, motorcycles, pedestrians, traffic lights, and traffic signs. These annotations include precise bounding boxes and class labels, enabling thorough evaluation of detection models across various environmental and lighting conditions.

\textbf{SHIFT}
The SHIFT\cite{sun2022shift} dataset is a synthetic benchmark specifically developed to evaluate the robustness of autonomous driving models under diverse and challenging conditions. A notable portion of the dataset consists of night-time scenes, making it particularly valuable for low-light object detection. SHIFT provides annotations for 8 object categories, including cars, trucks, pedestrians, and cyclists, along with dense labels and simulated environments featuring adverse weather and varying illumination.

\textbf{NuScenes}
The nuScenes\cite{caesar2020nuscenes} dataset is a large-scale benchmark designed for autonomous driving research, consisting of 1,000 driving scenes collected in diverse urban environments. Approximately 17\% of the scenes are captured during night-time, making it a valuable resource for assessing detection performance under low-light conditions. For the 3D object detection task, the dataset provides annotations for 10 object categories, including cars, trucks, buses, pedestrians, cyclists, motorcycles, construction vehicles, and traffic cones. Each annotated object includes its 3D bounding box, class label, and motion attributes, enabling comprehensive evaluations of detection models. As our task focuses on 2D detection labels, it is necessary to project the 3D bounding boxes onto the image plane. We use the images from the Front camera as the target projection view.

\begin{figure}
    \centering
    \includegraphics[scale=0.32]{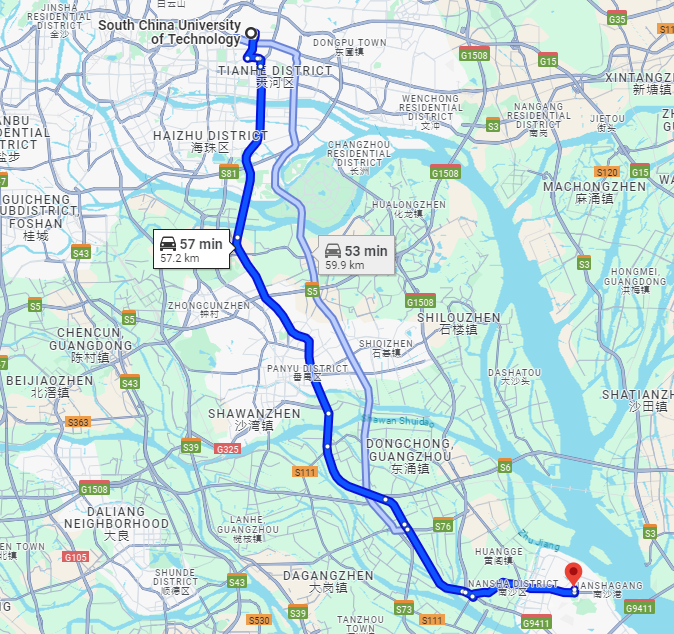}
    \caption{Data collection route 1: From the Wushan Campus of South China University of Technology to NANSHAGANG in Guangzhou\cite{googlemaps}. }
    \label{fig:route-G}
\end{figure}

\begin{figure}
    \centering
    \includegraphics[scale=0.28]{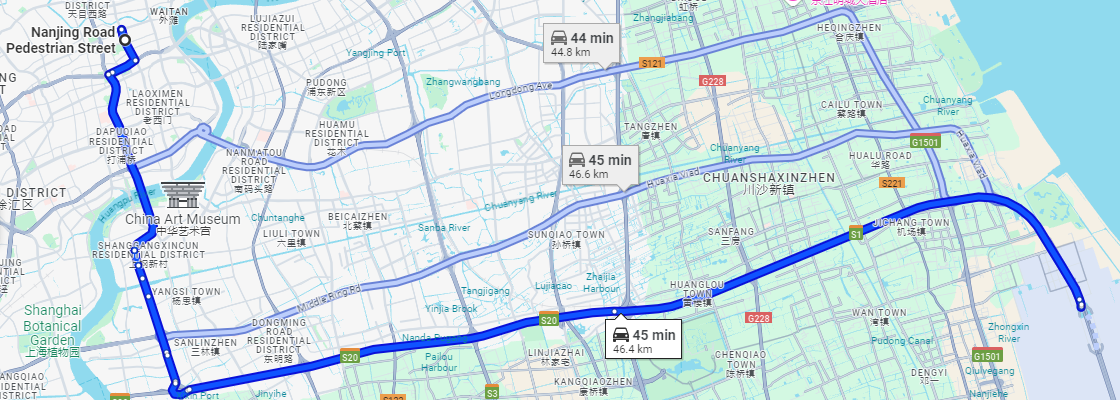}
    \caption{Data collection route 2: From Nanjing Road Pedestrain Street to hanghai Pudong International Airport in Shanghai\cite{googlemaps}. }
    \label{fig:route-S}
\end{figure}

\textbf{GSN}
To comprehensively evaluate the detection capabilities of the proposed WTEFNet method in low-light conditions, we constructed the GSN dataset, specifically designed for object detection in low-light road scenarios. The dataset comprises 2031 RGB images with a resolution of $1920\times1080$, including 685 images collected under rainy nighttime conditions. To enhance scene diversity, data were collected in two cities, Guangzhou and Shanghai. Figures \ref{fig:route-G} and \ref{fig:route-S} illustrate the complete collection routes.

\begin{figure}
    \centering
    \includegraphics[scale=0.35]{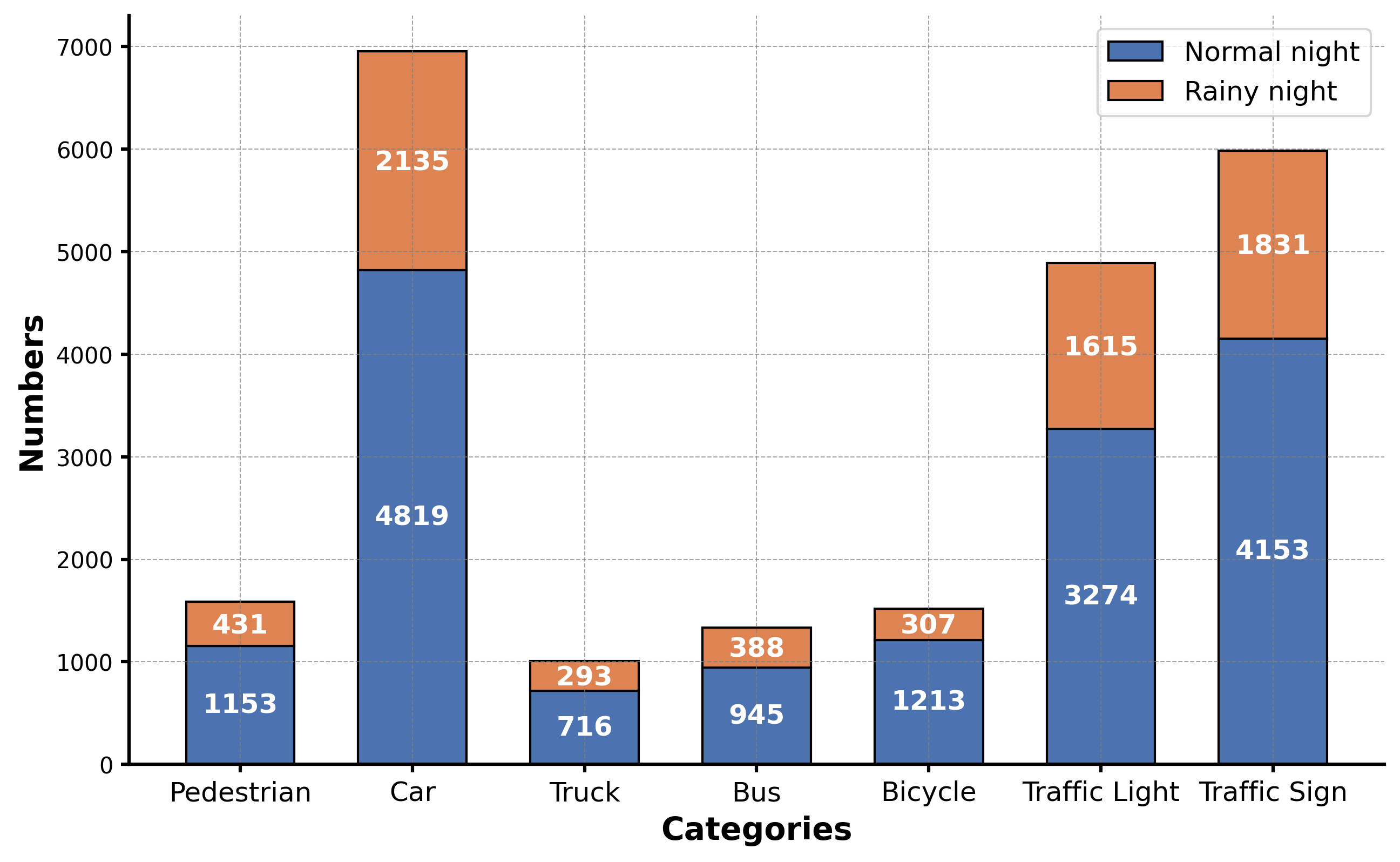}
    \caption{Bar Chart of Category Distribution in the GSN Dataset}
    \label{fig:data_distribution}
\end{figure}

Figure \ref{fig:data_distribution} illustrates the category distribution in the GSN dataset, which includes seven classes: pedestrian, car, truck, bus, bicycle, traffic light, and traffic sign. Blue and orange bars represent the number of samples under normal nighttime and rainy nighttime conditions, respectively, while white numbers indicate the exact sample count for each category.

\subsection{Evaluation Metrics}
Mean Average Precision (mAP) serves as a standard evaluation metric for object detection algorithms. It reflects the model’s overall accuracy in identifying and localizing objects across multiple categories. This metric is calculated by averaging precision scores over all object classes and various Intersection over Union (IoU) thresholds.

Precision (P) indicates the ratio of correctly identified positive instances to all predicted positive instances, while recall (R) denotes the proportion of correctly detected positives relative to the total number of actual positives. They\cite{al2021machine} are calculated as:
\begin{align}
P = \frac{TP}{TP + FP}, R = \frac{TP}{TP + FN} 
\end{align}
where $TP$, $FP$, and $FN$ denote the numbers of true positives, false positives, and false negatives, respectively.

The Average Precision (AP) for a single class is defined as the area under the Precision-Recall (PR) curve:

\begin{equation}
AP=\int_0^1P(R)dR
\end{equation}

The mAP is obtained by averaging the AP scores across all object classes. For $C$ categories:
\begin{equation}
mAP=\frac{1}{C}\sum_{c=1}^CAP_c
\end{equation}
Additionally, mAP can be evaluated under specific IoU thresholds, such as
mAP@IoU=0.5, mAP@IoU=0.75, or mAP@[0.5:0.95].

\subsection{Implementation Details}
We conduct training for 200 epochs on the BDD100K, SHIFT, nuScenes, and GSN datasets individually, utilizing Stochastic Gradient Descent (SGD) with a momentum of 0.912 and a learning rate set to 0.01. The batch size is fixed at 32. A pre-trained model\cite{ma2022toward} is employed for the low-light enhancement module to improve convergence and performance. Our network is implemented using the PyTorch framework, and all experiments are executed on a system equipped with four NVIDIA GeForce RTX 4090 GPUs.

\subsection{Experiment Results}
We evaluate our method against several advanced approaches with respect to object detection performance under low-light conditions. It should be emphasized that all evaluations are conducted exclusively on the low-light portions of each dataset. The comparison methods are divided into two primary categories. The first category includes object detection methods based on domain adaptation, such as DA F-RCNN\cite{chen2018domain}, UMT\cite{deng2021unbiased}, AT\cite{li2022cross}, WSA-YOLO\cite{hui2024wsa}, DAI-Net\cite{du2024boosting} and 2PCNet\cite{kennerley20232pcnet}. The second category encompasses object detection methods based on low-light enhancement, featuring representative model such as Zero-DCE\cite{li2021learning}, IAT\cite{cui2022you}, Retinexformer\cite{cai2023retinexformer}, UIA\cite{wang2024unsupervised}, FFENet\cite{liu2024ffenet}, and UPT-Flow\cite{xu2025upt}. It is worth noting that the latter methods were originally designed to enhance the clarity of dark images and enrich their visual features, which also contributes to improved object detection performance under low-light environments. In this experiment, low-light image enhancement networks are used as preprocessing modules for object detection. The details of all comparison methods are summarized in Table \ref{methods}.

\begin{figure*}[h]
    \centering
    \includegraphics[scale=0.75]{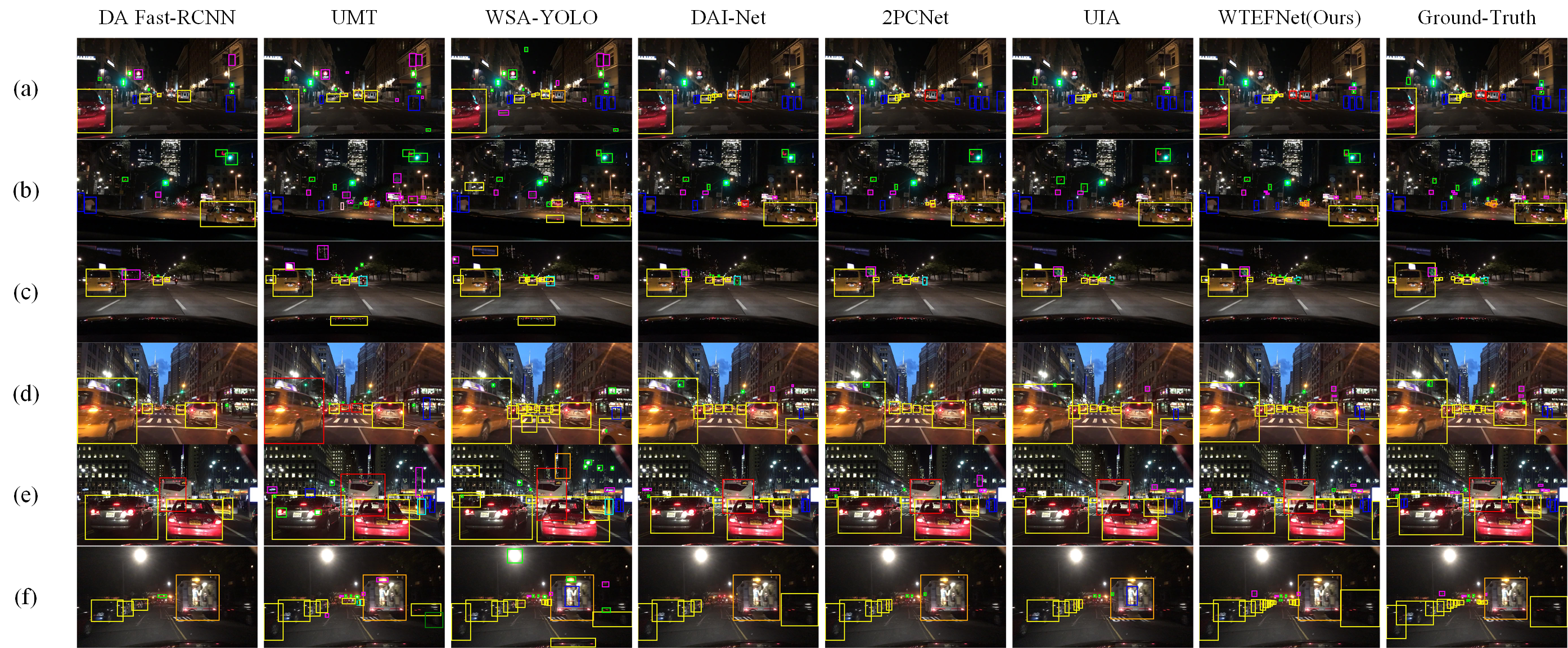}
    \caption{The detection results of seven methods on the BDD100K dataset are presented, with six representative images shown from top to bottom. From left to right, the columns correspond to the prediction results of DA Fast-RCNN, UMT, WSA-YOLO, DAI-Net, 2PCNet, UIA, WTEFNet (Ours), and the ground truth, respectively.}
    \label{fig:visual_BDD100K}
\end{figure*}

\begin{figure*}[h]
    \centering
    \includegraphics[scale=0.75]{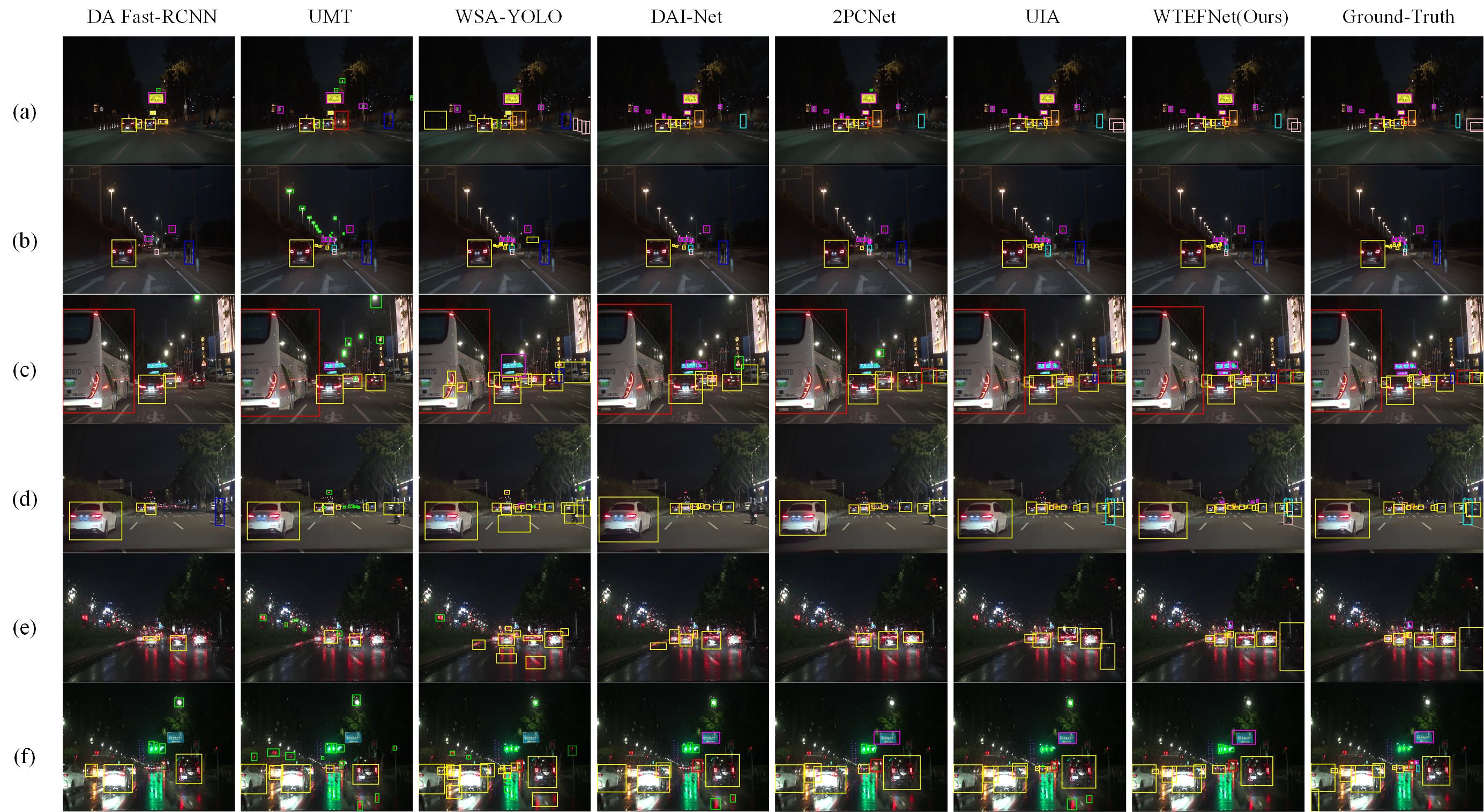}
    \caption{The detection results of seven methods on the GSN dataset are presented, with six representative images shown from top to bottom.}
    \label{fig:visual_GSN}
\end{figure*}

\textbf{BDD100K}
Table \ref{bdd100k_results} presents the quantitative results of all methods on the BDD100K dataset, where blue numbers indicate the best performance and green numbers represent the second-best results. Since trains do not appear in typical urban road scenarios, the "train" category was removed during training and evaluation. Among domain adaptation-based object detection methods, 2PCNet achieved the highest detection performance. For low-light enhancement-based object detection methods, we selected YOLOv10 and YOLOv12 as baseline detectors. Overall, the WTEFNet framework achieved the best result of mAP@0.5 = 49.0 when paired with YOLOv12, while it achieved the second-best result of mAP@0.5 = 48.9 when paired with YOLOv10. Further analysis shows that YOLOv12 performs better in detecting medium and large objects, while YOLOv10 excels at detecting small objects.

\begin{table}[h]
\centering
\caption{Comparison Methods}
\label{methods}
\setlength{\tabcolsep}{8pt} 
\renewcommand{\arraystretch}{1.2} 
\begin{tabular}{ccc}
\hline
\rule{0pt}{10pt} 
\raisebox{1pt}{\textbf{Method}} & \raisebox{1pt}{\textbf{Year}} & \raisebox{1pt}{\textbf{Conference/Journal}} \\ \hline
DA Fast-RCNN\cite{chen2018domain}    & 2018          & CVPR                        \\
UMT\cite{deng2021unbiased}             & 2021          & CVPR                        \\
AT\cite{li2022cross}              & 2022          & CVPR                        \\
2PCNet\cite{kennerley20232pcnet}          & 2023          & CVPR                        \\
WSA-YOLO\cite{hui2024wsa}        & 2024          & TIM                        \\
DAI-Net\cite{du2024boosting}         & 2024          & CVPR                        \\
Zero-DCE\cite{li2021learning}        & 2021          & TPAMI                       \\
IAT\cite{cui2022you}             & 2022          & BMVC                        \\
Retinexformer\cite{cai2023retinexformer}   & 2023          & ICCV                        \\
FFENet\cite{liu2024ffenet}          & 2024          & TIM                         \\
UIA\cite{wang2024unsupervised}         & 2024          & TPAMI                       \\
UPT-Flow\cite{xu2025upt}        & 2025          & PR                       \\ \hline
\end{tabular}
\end{table}

\begin{table*}[h]
\centering
\caption{Results on BDD100K Dataset  }
\label{bdd100k_results}
\renewcommand{\arraystretch}{1.2}
\setlength{\tabcolsep}{4.5pt} 
\begin{tabular}{c|cccccccccc}
\hline
\rule{0pt}{10pt} 
\raisebox{1pt}{\textbf{Method}}    & \raisebox{1pt}{{\color[HTML]{000000} AP@0.5}}        & \raisebox{1pt}{Pedestrian}         & \raisebox{1pt}{Rider}                                & \raisebox{1pt}{Car}                & \raisebox{1pt}{Truck}                                & \raisebox{1pt}{Bus}                & \raisebox{1pt}{Motorcycle}                           & \raisebox{1pt}{Bicycle}            & \raisebox{1pt}{Traffic Light}                        & \raisebox{1pt}{Traffic Sign}                         \\ \hline
DA Fast-RCNN\cite{chen2018domain}          & 41.3                                 & 50.4                                 & 30.3                                 & 66.3                                 & 46.8                                 & 48.3                                 & 32.6                                 & 41.4                                 & 41.0                                 & 56.2                                 \\
AT\cite{li2022cross}                    & 38.5                                 & 42.3                                 & 30.4                                 & 60.8                                 & 48.9                                 & 52.1                                 & 34.5                                 & 42.7                                 & 29.1                                 & 43.9                                 \\
UMT\cite{deng2021unbiased}                   & 36.2                                 & 46.5                                 & 26.1                                 & 46.8                                 & 44.0                                 & 46.3                                 & 28.2                                 & 40.2                                 & 31.6                                 & 52.7                                 \\
2PCNet\cite{kennerley20232pcnet}                & 46.4                                 & 54.4                                 & 30.8                                 & 73.1                                 & 53.8                                 & 55.2                                 & 37.5                                 & 44.5                                 & 49.4                                 & 65.2                                 \\
WSA-YOLO\cite{hui2024wsa}                & 44.7                                 & 53.3                                 & 28.5                                 & 68.9                                 &51.1                                  & 51.8                                 & 36.2                                 & 41.7                                 & 46.4                                 & 62.5                                 \\
DAI-Net\cite{du2024boosting}                & 44.3                                 & 52.1                                 & 27.8                                 & 69.4                                 & 51.8                                 & 53.7                                 & 35.6                                 & 41.5                                 & 44.6                                 & 61.2                                 \\
Zero-DCE\cite{li2021learning}+YOLOv10\cite{wang2024yolov10}           & 44.1                                 & 53.8                                 & 28.3                                 & 68.4                                 & 52.5                                 & 51.4                                 & 33.8                                 & 41.4                                 & 42.6                                 & 59.1                                 \\
Zero-DCE\cite{li2021learning}+YOLOv12\cite{tian2025yolov12}          & 43.6                                 & 52.4                                 & 28.2                                 & 69.5                                 & 52.9                                 & 52.2                                 & 33.6                                 & 41.5                                 & 41.9                                 & 58.8                                 \\
IAT\cite{cui2022you}+YOLOv10\cite{wang2024yolov10}           & 43.4                                 & 53.6                                 & 28.7                                 & 68.5                                 & 50.9                                 & 50.7                                 & 32.6                                 & 41.9                                 & 42.1                                 & 59.8                                 \\
IAT\cite{cui2022you}+YOLOv12\cite{tian2025yolov12}          & 43.9                                 & 54.2                                 & 28.2                                 & 69.2                                 & 51.6                                 & 51.1                                 & 33.9                                 & 42.2                                 & 41.8                                 & 60.3                                 \\
Retinexformer\cite{cai2023retinexformer}+YOLOv10\cite{wang2024yolov10}            & 45.1                                 & 55.4                                 & 32.2                                 & 71.5                                 & 51.3                                 & 54.5                                 & 36.3                                 & 43.4                                 & 44.5                                 & 61.2                                 \\
Retinexformer\cite{cai2023retinexformer}+YOLOv12\cite{tian2025yolov12}           & 45.0                                 & 55.2                                 & 31.7                                 & 72.2                                 & 52.4                                 & 55.1                                 & 35.9                                 & 43.2                                 & 44.3                                 & 61.0                                 \\
FFENet\cite{liu2024ffenet}+YOLOv10\cite{wang2024yolov10}            & 44.2                                 & 53.2                                 & 29.4                                 & 68.5                                 & 51.1                                 & 54.3                                 & 33.9                                 & 42.7                                 & 43.4                                 & 60.5                                 \\
FFENet\cite{liu2024ffenet}+YOLOv12\cite{tian2025yolov12}           & 44.5                                 & 51.9                                 & 29.1                                 & 70.4                                 & 52.8                                 & 55.2                                 & 33.5                                 & 43.0                                 & 42.5                                 & 60.3                                 \\
UIA\cite{wang2024unsupervised}+YOLOv10\cite{wang2024yolov10}  & 45.8                                 & 55.8                                 & 31.1                                 & 73.1                                 & 53.3                                 & 55.3                                 & 36.4                                 & 44.1                                 & 46.9                                 & 63.0                                 \\
UIA\cite{wang2024unsupervised}+YOLOv12\cite{tian2025yolov12} & 46.1                                 & 55.6                                 & 30.9                                 & 73.5                                 & 53.8                                 & 56.5                                 & 36.4                                 & 45.0                                 & 46.5                                 & 63.2                                 \\
UPT-Flow\cite{xu2025upt}+YOLOv10\cite{wang2024yolov10}       & 44.9                                 & 55.7                                 & 29.5                                 & 69.3                                 & 51.9                                 & 54.9                                 & 34.1                                 & 42.8                                 & 44.7                                 & 62.1                                 \\
UPT-Flow\cite{xu2025upt}+YOLOv12\cite{tian2025yolov12}      & 45.4                                 & 55.3                                 & 28.8                                 & 72.8                                 & 52.4                                 & 55.3                                 & 33.6                                 & 42.9                                 & 43.3                                 & 61.4                                 \\
\rowcolor{blue!8} WTEFNet(Ours)+YOLOv10\cite{wang2024yolov10}    & {\color{green} \textbf{48.9}} & {\color{blue} \textbf{57.4}} & {\color{blue} \textbf{32.5}} & {\color{green} \textbf{75.8}} & {\color{green} \textbf{54.9}} & {\color{green} \textbf{57.3}} & {\color{green} \textbf{38.7}} & {\color{green} \textbf{46.7}} & {\color{blue} \textbf{51.6}} & {\color{blue} \textbf{66.2}} \\
\rowcolor{blue!8} WTEFNet(Ours)+YOLOv12\cite{tian2025yolov12}   & {\color{blue} \textbf{49.0}} & {\color{green} \textbf{57.1}} & {\color{green} \textbf{32.3}} & {\color{blue} \textbf{76.0}} & {\color{blue} \textbf{55.7}} & {\color{blue} \textbf{58.4}} & {\color{blue} \textbf{38.8}} & {\color{blue} \textbf{46.9}} & {\color{green} \textbf{50.9}} & {\color{green} \textbf{65.8}} \\ \hline
\end{tabular}
\end{table*}

\begin{table*}[h]
\centering
\caption{Results on SHIFT Dataset  }
\label{shift_results}
\renewcommand{\arraystretch}{1.2}
\begin{tabular}{c|ccccccc}
\hline
\rule{0pt}{10pt} 
\raisebox{1pt}{\textbf{Method}}     & \raisebox{1pt}{{\color[HTML]{000000} AP@0.5}}     & \raisebox{1pt}{Pedestrian}    & \raisebox{1pt}{Car}                & \raisebox{1pt}{Truck}                                & \raisebox{1pt}{Bus}                & \raisebox{1pt}{Motorcycle}                           & \raisebox{1pt}{Bicycle}                   \\ \hline
DA Fast-RCNN\cite{chen2018domain}          & 43.7                                 & 43.0                                 & 48.8                                 & 47.8                                 & 52.1                                 & 19.9                                 & 55.8                                 \\
AT\cite{li2022cross}                    & 38.9                                 & 25.8                                 & 33.0                                 & 54.7                                 & 49.5                                 & 20.7                                 & 52.3                                 \\
UMT\cite{deng2021unbiased}                   & 31.1                                 & 7.7                                  & 47.5                                 & 18.4                                 & 46.8                                 & 16.6                                 & 49.2                                 \\
2PCNet\cite{kennerley20232pcnet}                & 49.1                                 & 51.4                                 & 54.6                                 & 54.8                                 & 56.6                                 & 23.9                                 & 54.2                                 \\
WSA-YOLO\cite{hui2024wsa}                & 45.4                                 & 47.0                                 & 50.6                                 & 50.4                                 & 53.1                                 & 20.5                                 & 50.9                                 \\
DAI-Net\cite{du2024boosting}                & 45.2                                 & 46.5                                 & 51.7                                 & 50.8                                 & 52.4                                 & 19.6                                 & 50.3                                 \\
Zero-DCE\cite{li2021learning}+YOLOv10\cite{wang2024yolov10}           & 46.7                                 & 48.3                                 & 52.9                                 & 51.8                                 & 53.3                                 & 21.1                                 & 52.6                                 \\
Zero-DCE\cite{li2021learning}+YOLOv12\cite{tian2025yolov12}          & 46.4                                 & 47.2                                 & 53.3                                 & 51.1                                 & 53.6                                 & 20.7                                 & 52.2                                 \\
IAT\cite{cui2022you}+YOLOv10\cite{wang2024yolov10}           & 46.5                                 & 47.5                                 & 52.6                                 & 51.4                                 & 52.7                                 & 21.6                                 & 52.9                                 \\
IAT\cite{cui2022you}+YOLOv12\cite{tian2025yolov12}          & 46.3                                 & 47.3                                 & 53.1                                 & 50.7                                 & 53.0                                 & 21.2                                 & 52.4                                 \\
Retinexformer\cite{cai2023retinexformer}+YOLOv10\cite{wang2024yolov10}            & 47.6                                 & 48.4                                 & 53.6                                 & 52.9                                 & 52.9                                 & 22.9                                 & 54.6                                 \\
Retinexformer\cite{cai2023retinexformer}+YOLOv12\cite{tian2025yolov12}           & 47.4                                 & 48.1                                 & 54.4                                 & 52.5                                 & 53.0                                 & 21.7                                 & 54.4                                 \\
FFENet\cite{liu2024ffenet}+YOLOv10\cite{wang2024yolov10}            & 45.6                                 & 47.2                                 & 52.7                                 & 49.5                                 & 52.4                                 & 20.4                                 & 51.1                                 \\
FFENet\cite{liu2024ffenet}+YOLOv12\cite{tian2025yolov12}           & 45.7                                 & 47.4                                 & 53.4                                 & 49.9                                 & 52.7                                 & 20.2                                 & 50.8                                 \\
UIA\cite{wang2024unsupervised}+YOLOv10\cite{wang2024yolov10}  & 47.8                                 & 48.9                                 & 54.6                                 & 52.5                                 & 53.8                                 & 22.7                                 & 54.1                                 \\
UIA\cite{wang2024unsupervised}+YOLOv12\cite{tian2025yolov12} & 47.6                                 & 49.1                                 & 54.8                                 & 53.1                                 & 52.7                                 & 22.4                                 & 53.6                                 \\
UPT-Flow\cite{xu2025upt}+YOLOv10\cite{wang2024yolov10}       & 47.0                                 & 48.8                                 & 53.5                                 & 52.4                                 & 52.1                                 & 22.2                                 & 53.2                                 \\
UPT-Flow\cite{xu2025upt}+YOLOv12\cite{tian2025yolov12}      & 46.9                                 & 48.5                                 & 53.3                                 & 52.8                                 & 53.2                                 & 21.3                                 & 52.4                                 \\
\rowcolor{blue!8} WTEFNet(Ours)+YOLOv10\cite{wang2024yolov10}    & {\color{blue} \textbf{50.3}} & {\color{blue} \textbf{53.3}} & {\color{blue} \textbf{55.7}} & {\color{green} \textbf{55.3}} & {\color{green} \textbf{57.0}} & {\color{blue} \textbf{24.4}} & {\color{blue} \textbf{56.2}} \\
\rowcolor{blue!8} WTEFNet(Ours)+YOLOv12\cite{tian2025yolov12}   & {\color{green} \textbf{49.8}} & {\color{green} \textbf{52.1}} & {\color{green} \textbf{55.2}} & {\color{blue} \textbf{55.9}} & {\color{blue} \textbf{57.3}} & {\color{green} \textbf{23.4}} & {\color{green} \textbf{54.8}} \\ \hline
\end{tabular}
\end{table*}

\begin{table*}[h]
\centering
\caption{Results on nuScenes Dataset  }
\label{NuScenes_results}
\renewcommand{\arraystretch}{1.2}
\begin{tabular}{c|ccccccc}
\hline
\rule{0pt}{10pt} 
\raisebox{1pt}{Method}  & {\color[HTML]{000000} AP@0.5}       &\raisebox{1pt}{Pedestrian}                           &\raisebox{1pt}{ Car}                                &\raisebox{1pt} {Truck}         &\raisebox{1pt} {Motorcycle}          &\raisebox{1pt} {Bicycle}                             &\raisebox{1pt}{Barrier}                              \\ \hline
DA Fast-RCNN\cite{chen2018domain}          & 34.0                                 & 35.7                                 & 44.5                                 & 34.1                                 & 25.5                                 & 26.2                                 & 38.3                                 \\
AT\cite{li2022cross}                    & 26.6                                 & 21.8                                 & 33.2                                 & 26.4                                 & 24.7                                 & 20.9                                 & 32.5                                 \\
UMT\cite{deng2021unbiased}                   & 26.0                                 & 27.5                                 & 31.5                                 & 28.8                                 & 20.3                                 & 17.3                                 & 30.4                                 \\
2PCNet\cite{kennerley20232pcnet}                & 36.8                                 & 36.2                                 & 47.8                                 & 37.2                                 & 31.7                                 & 30.5                                 & 37.3                                 \\
WSA-YOLO\cite{hui2024wsa}                & 37.5                                 & 36.9                                 & 48.4                                 & 37.7                                 & 32.6                                & 31.2                                 & 38.3                                 \\
DAI-Net\cite{du2024boosting}                & 35.6                                 & 34.9                                 & 48.1                                 & 35.5                                 & 29.4                                 & 30.3                                 & 35.2                                 \\
Zero-DCE\cite{li2021learning}+YOLOv10\cite{wang2024yolov10}           & 37.6                                 & 37.5                                 & 53.2                                 & 34.5                                 & 32.4                                 & 28.8                                 & 38.9                                 \\
Zero-DCE\cite{li2021learning}+YOLOv12\cite{tian2025yolov12}          & 37.9                                 & 38.1                                 & 54.8                                 & 36.0                                 & 31.9                                 & 28.5                                 & 38.4                                 \\
IAT\cite{cui2022you}+YOLOv10\cite{wang2024yolov10}           & 37.7                                 & 38.4                                 & 52.9                                 & 34.7                                 & 31.6                                 & 30.9                                 & 37.8                                 \\
IAT\cite{cui2022you}+YOLOv12\cite{tian2025yolov12}          & 37.9                                 & 38.9                                 & 54.3                                 & 35.7                                 & 31.2                                 & 29.8                                 & 37.6                                 \\
Retinexformer\cite{cai2023retinexformer}+YOLOv10\cite{wang2024yolov10}            & 38.9                                 & 38.5                                 & 55.1                                 & 35.8                                 & 31.4                                 & 31.7                                 & 40.6                                 \\
Retinexformer\cite{cai2023retinexformer}+YOLOv12\cite{tian2025yolov12}           & 38.6                                 & 37.3                                 & 55.8                                 & 36.4                                 & 32.1                                 & 30.5                                 & 39.7                                 \\
FFENet\cite{liu2024ffenet}+YOLOv10\cite{wang2024yolov10}            & 38.4                                 & 38.2                                 & 53.9                                 & 35.1                                 & 31.5                                 & 31.3                                 & 40.9                                 \\
FFENet\cite{liu2024ffenet}+YOLOv12\cite{tian2025yolov12}           & 38.1                                 & 37.9                                 & 54.4                                 & 36.2                                 & 30.5                                 & 29.6                                 & 40.2                                 \\
UIA\cite{wang2024unsupervised}+YOLOv10\cite{wang2024yolov10}  & 39.6                                 & 39.3                                 & 57.2                                 & 36.3                                 & 32.5                                 & 31.4                                 & 40.8                                 \\
UIA\cite{wang2024unsupervised}+YOLOv12\cite{tian2025yolov12} & 39.7                                 & 39.5                                 & 58.2                                 & 37.0                                 & 32.2                                 & 30.7                                 & 40.5                                 \\
UPT-Flow\cite{xu2025upt}+YOLOv10\cite{wang2024yolov10}       & 36.1                                 & 36.7                                 & 50.3                                 & 34.4                                 & 30.3                                 & 28.6                                 & 36.3                                 \\
UPT-Flow\cite{xu2025upt}+YOLOv12\cite{tian2025yolov12}      & 36.3                                 & 37.3                                 & 51.6                                 & 34.9                                 & 29.8                                 & 28.5                                 & 35.7                                 \\
\rowcolor{blue!8} WTEFNet(Ours)+YOLOv10\cite{wang2024yolov10}    & {\color{green} \textbf{41.6}} & {\color{blue} \textbf{40.6}} & {\color{green} \textbf{61.5}} & {\color{green} \textbf{40.2}} & {\color{blue} \textbf{34.1}} & {\color{blue} \textbf{31.5}} & {\color{blue} \textbf{41.7}} \\
\rowcolor{blue!8} WTEFNet(Ours)+YOLOv12\cite{tian2025yolov12}   & {\color{blue} \textbf{41.8}} & {\color{green} \textbf{40.2}} & {\color{blue} \textbf{62.6}} & {\color{blue} \textbf{41.9}} & {\color{green} \textbf{33.8}} & {\color{green} \textbf{30.9}} & {\color{green} \textbf{41.4}} \\ \hline
\end{tabular}
\end{table*}

\begin{table*}[h]
\centering
\caption{Results on GSN Dataset  }
\label{GSN_results}
\footnotesize
\renewcommand{\arraystretch}{1.2} 
\setlength{\tabcolsep}{2.6pt} 
\begin{tabular}{c|cc|cc|cc|cc|cc|cc|cc|cc}
\hline
                                  & \multicolumn{2}{c|}{{\color[HTML]{333333} AP@0.5}}                          & \multicolumn{2}{c|}{{\color[HTML]{333333} Pedestrian}}                      & \multicolumn{2}{c|}{Car}                                                    & \multicolumn{2}{c|}{Truck}                                                  & \multicolumn{2}{c|}{Bus}                                                    & \multicolumn{2}{c|}{Bicycle}                                                & \multicolumn{2}{c|}{Traffic Light}                                          & \multicolumn{2}{c}{Traffic Sign}                                            \\ \cline{2-17} 
\multirow{-2}{*}{\textbf{Method}} & rain                                 & no rain                              & rain                                 & no rain                              & rain                                 & no rain                              & rain                                 & no rain                              & rain                                 & no rain                              & rain                                 & no rain                              & rain                                 & no rain                              & rain                                 & no rain                              \\ \hline
DA Fast-RCNN\cite{chen2018domain}                      & 38.5                                 & 45.2                                 & 37.4                                 & 44.5                                 & 57.6                                 & 64.7                                 & 37.5                                 & 46.2                                 & 39.8                                 & 44.9                                 & 25.4                                 & 30.5                                 & 30.3                                 & 35.3                                 & 41.2                                 & 50.4                                 \\
AT\cite{li2022cross}                                & {\color[HTML]{333333} 36.2}          & {\color[HTML]{333333} 43.2}          & {\color[HTML]{333333} 33.8}          & {\color[HTML]{333333} 37.9}          & {\color[HTML]{333333} 47.3}          & {\color[HTML]{333333} 63.4}          & 36.4                                 & 48.3                                 & 43.1                                 & 49.6                                 & 28.5                                 & 31.2                                 & 24.2                                 & 25.5                                 & 40.4                                 & 46.5                                 \\
UMT\cite{deng2021unbiased}                               & 34.7                                 & 37.0                                 & 35.3                                 & 39.8                                 & 40.1                                 & 45.2                                 & 38.2                                 & 39.7                                 & 39.9                                 & 40.6                                 & 27.5                                 & 29.3                                 & 19.9                                 & 22.0                                 & 41.7                                 & 42.1                                 \\
2PCNet\cite{kennerley20232pcnet}                            & 40.5                                 & 48.1                                 & 37.8                                 & 47.9                                 & 60.3                                 & 70.5                                 & 41.4                                 & 48.5                                 & 41.3                                 & 46.6                                 & 28.0                                 & 31.5                                 & 31.2                                 & 36.1                                 & 43.4                                 & 55.4                                 \\
WSA-YOLO\cite{hui2024wsa}                            & 41.0                                 & 48.5                                 & 38.5                                 & 48.5                                 & 61.4                                 & 69.9                                 & 41.9                                 & 49.1                                 & 42.2                                 & 46.8                                 & 28.8                                 & 33.4                                 & 30.6                                 & 37.5                                 & 43.7                                 & 54.6                                 \\
DAI-Net\cite{du2024boosting}                            & 40.9                                 & 47.6                                 & 38.2                                 & 47.5                                 & 59.7                                 & 69.4                                 & 41.8                                 & 47.6                                 & 41.7                                 & 47.2                                 & 28.4                                 & 32.3                                 & 31.3                                 & 35.8                                 & 44.1                                 & 53.1                                 \\
Zero-DCE\cite{li2021learning}+YOLOv10\cite{wang2024yolov10}                       & 40.8                                 & 48.7                                 & 38.1                                 & 47.6                                 & 59.9                                 & 68.4                                 & 40.4                                 & 49.5                                 & 42.6                                 & 48.3                                 & 28.5                                 & 34.7                                 & 32.1                                 & 37.2                                 & 43.7                                 & 55.3                                 \\
Zero-DCE\cite{li2021learning}+YOLOv12\cite{tian2025yolov12}                      & 40.4                                 & 48.3                                 & 37.5                                 & 47.2                                 & 59.1                                 & 68.6                                 & 41.3                                 & 49.9                                 & 42.8                                 & 49.2                                 & 27.7                                 & 32.5                                 & 31.4                                 & 36.4                                 & 43.1                                 & 54.5                                 \\
IAT\cite{cui2022you}+YOLOv10\cite{wang2024yolov10}                       & 41.0                                 & 48.5                                 & 39.2                                 & 48.1                                 & 59.4                                 & 69.2                                 & 40.9                                 & 49.8                                 & 42.3                                 & 47.5                                 & 28.7                                 & 33.1                                 & 32.5                                 & 38.9                                 & 44.4                                 & 53.2                                 \\
IAT\cite{cui2022you}+YOLOv12\cite{tian2025yolov12}                      & 40.6                                 & 47.9                                 & 37.7                                 & 47.5                                 & 58.8                                 & 69.6                                 & 41.5                                 & 50.2                                 & 42.6                                 & 48.1                                 & 27.9                                 & 31.4                                 & 32.2                                 & 35.6                                 & 43.6                                 & 52.9                                 \\
Retinexformer\cite{cai2023retinexformer}+YOLOv10\cite{wang2024yolov10}                        & 43.4                                 & 49.9                                 & 39.1                                 & 49.5                                 & 61.6                                 & 70.2                                 & 43.8                                 & 48.3                                 & 45.4                                 & 49.1                                 & 30.6                                 & 35.9                                 & 38.9                                 & 40.4                                 & 44.7                                 & 56.2                                 \\
Retinexformer\cite{cai2023retinexformer}+YOLOv12\cite{tian2025yolov12}                       & 43.6                                 & 49.3                                 & 39.4                                 & 48.1                                 & 61.9                                 & 69.8                                 & 44.0                                 & 47.5                                 & 46.2                                 & 50.5                                 & 30.3                                 & 35.5                                 & 38.5                                 & 38.5                                 & 44.6                                 & 55.1                                 \\
FFENet\cite{liu2024ffenet}+YOLOv10\cite{wang2024yolov10}                        & 41.4                                 & 49.5                                 & 38.3                                 & 48.5                                 & 59.2                                 & 68.3                                 & 41.7                                 & 49.4                                 & 43.9                                 & 50.8                                 & 29.1                                 & 35.9                                 & 33.2                                 & 37.7                                 & 44.1                                 & 55.6                                 \\
FFENet\cite{liu2024ffenet}+YOLOv12\cite{tian2025yolov12}                       & 41.2                                 & 49.6                                 & 37.9                                 & 48.7                                 & 59.4                                 & 69.4                                 & 41.2                                 & 50.1                                 & 44.5                                 & 51.3                                 & 28.8                                 & 35.3                                 & 32.9                                 & 37.5                                 & 43.5                                 & 54.8                                 \\
UIA\cite{wang2024unsupervised}+YOLOv10\cite{wang2024yolov10}              & 43.3                                 & 50.5                                 & 39.7                                 & 50.1                                 & 60.5                                 & 70.0                                 & 44.6                                 & 49.2                                 & 44.9                                 & 49.7                                 & 30.4                                 & 37.2                                 & 39.4                                 & 39.5                                 & 43.9                                 & 56.8                                 \\
UIA\cite{wang2024unsupervised}+YOLOv12\cite{tian2025yolov12}             & 43.1                                 & 50.1                                 & 39.5                                 & 49.4                                 & 62.4                                 & 70.6                                 & 43.4                                 & 49.5                                 & 45.5                                 & 50.7                                 & 29.5                                 & 36.3                                 & 38.2                                 & 38.8                                 & 43.2                                 & 55.2                                 \\
UPT-Flow\cite{xu2025upt}+YOLOv10\cite{wang2024yolov10}                   & 42.6                                 & 50.2                                 & 40.8                                 & 49.8                                 & 60.6                                 & 70.1                                 & 42.4                                 & 49.6                                 & 44.3                                 & 51.2                                 & 29.9                                 & 36.7                                 & 34.7                                 & 38.4                                 & 45.7                                 & 55.9                                 \\
UPT-Flow\cite{xu2025upt}+YOLOv12\cite{tian2025yolov12}                  & 42.1                                 & 49.8                                 & 40.1                                 & 49.2                                 & 61.5                                 & 70.5                                 & 42.7                                 & 49.7                                 & 44.5                                 & 51.4                                 & 28.5                                 & 35.4                                 & 33.9                                 & 37.9                                 & 43.6                                 & 54.4                                 \\
\rowcolor{blue!8} WTEFNet(Ours)+YOLOv10\cite{wang2024yolov10}            & {\color{blue} \textbf{45.8}} & {\color{blue} \textbf{53.5}} & {\color{blue} \textbf{43.7}} & {\color{blue} \textbf{53.9}} & {\color{blue} \textbf{63.5}} & {\color{blue} \textbf{72.7}} & {\color{green} \textbf{47.3}} & {\color{green} \textbf{51.8}} & {\color{green} \textbf{46.2}} & {\color{green} \textbf{52.5}} & {\color{blue} \textbf{33.4}} & {\color{blue} \textbf{39.7}} & {\color{blue} \textbf{41.2}} & {\color{blue} \textbf{44.2}} & {\color{blue} \textbf{45.5}} & {\color{blue} \textbf{59.7}} \\
\rowcolor{blue!8} WTEFNet(Ours)+YOLOv12\cite{tian2025yolov12}            & {\color{green} \textbf{45.5}} & {\color{green} \textbf{52.8}} & {\color{green} \textbf{42.9}} & {\color{green} \textbf{53.3}} & {\color{green} \textbf{63.2}} & {\color{green} \textbf{71.9}} & {\color{blue} \textbf{47.7}} & {\color{blue} \textbf{52.2}} & {\color{blue} \textbf{46.4}} & {\color{blue} \textbf{52.8}} & {\color{green} \textbf{32.9}} & {\color{green} \textbf{38.5}} & {\color{green} \textbf{40.8}} & {\color{green} \textbf{41.7}} & {\color{green} \textbf{44.9}} & {\color{green} \textbf{58.9}} \\ \hline
\end{tabular}

\end{table*}

\textbf{SHIFT}
Table \ref{shift_results} shows the detection results of different methods on the SHIFT dataset. As illustrated in the table, WTEFNet consistently achieves the best detection performance across all categories. When paired with YOLOv10 and YOLOv12, it attains mAP@0.5 = 50.3 and mAP@0.5 = 49.8, respectively, with notable improvements in the Pedestrian and Bicycle categories. Compared to the second-best 2PCNet, WTEFNet improves detection performance in the Pedestrian and Bicycle categories by 1.9 and 1.8 percentage points, respectively.

\textbf{NuScenes}
Table \ref{NuScenes_results} summarizes the experimental results on the nuScenes dataset. In general, compared to the BDD100K and SHIFT datasets, detection performance on nuScenes is lower across all methods. This is primarily due to severely low illumination in nighttime images, resulting in fewer distinguishable features. However, despite these challenges, WTEFNet still achieves the best detection results among all evaluated methods. Specifically, when using YOLOv10 as the detector, WTEFNet outperforms the second-best UIA method by 2.5\%. Similarly, when using YOLOv10, WTEFNet surpasses UIA by 2.2\%. These findings demonstrate that the proposed WTEFNet framework can effectively handle extremely low-light scenarios to a certain extent.

\begin{table*}[h]
\centering
\caption{Comparison with Different Advanced Object Detectors on BDD100K, NuScenes and GSN Datasets}
\label{generalization_evaluation}
\setlength{\tabcolsep}{1.3pt} 
\renewcommand{\arraystretch}{1.3} 
\begin{tabular}{c|c|ccc|ccc|ccc|ccc}
\hline
                                  &                                                                          & \multicolumn{3}{c|}{BDD100K}                                                                        & \multicolumn{3}{c|}{NuScenes}                                                                       & \multicolumn{3}{c|}{GSN(no rain)}                                                                   & \multicolumn{3}{c}{GSN(rain)}                                                                       \\ \cline{3-14} 
\multirow{-2}{*}{\textbf{Method}} & \multirow{-2}{*}{\begin{tabular}[c]{@{}c@{}}WTE-\\   FNet\end{tabular}} & {\color[HTML]{333333} AP@0.5:0.95} & {\color[HTML]{333333} AP@0.5} & {\color[HTML]{333333} AP@0.75} & {\color[HTML]{333333} AP@0.5:0.95} & {\color[HTML]{333333} AP@0.5} & {\color[HTML]{333333} AP@0.75} & {\color[HTML]{333333} AP@0.5:0.95} & {\color[HTML]{333333} AP@0.5} & {\color[HTML]{333333} AP@0.75} & {\color[HTML]{333333} AP@0.5:0.95} & {\color[HTML]{333333} AP@0.5} & {\color[HTML]{333333} AP@0.75} \\ \hline
                                  & \checkmark                                                               & 31.4                               & 48.9                          & 31.8                           & 22.5                               & 41.6                          & 23.3                           & 31.9                               & 53.3                          & 32.4                           & 27.8                               & 45.8                          & 27.3                           \\
\multirow{-2}{*}{\begin{tabular}[c]{@{}c@{}}YOLO-\\ v10\cite{wang2024yolov10}\end{tabular}}          & $\times$                                                               & \cellcolor{blue!8}28.4                               & \cellcolor{blue!8}43.3                          & \cellcolor{blue!8}27.7                           & \cellcolor{blue!8}20.3                               & \cellcolor{blue!8}36.3                          & \cellcolor{blue!8}20.7                           & \cellcolor{blue!8}27.9                               & \cellcolor{blue!8}47.8                          & \cellcolor{blue!8}28.3                           & \cellcolor{blue!8}24.9                               & \cellcolor{blue!8}40.5                          & \cellcolor{blue!8}24.4                           \\ \hline
                                  & \checkmark                                                               & 31.2                               & 49.0                          & 31.9                           & 24.7                               & 41.8                          & 23.6                           & 32.6                               & 52.8                          & 32.1                           & 27.7                               & 45.5                          & 27.0                           \\
\multirow{-2}{*}{\begin{tabular}[c]{@{}c@{}}YOLO-\\ v12\cite{tian2025yolov12}\end{tabular}}         & $\times$                                                               & \cellcolor{blue!8}27.7                               & \cellcolor{blue!8}43.7                          & \cellcolor{blue!8}28.1                           & \cellcolor{blue!8}21.8                               & \cellcolor{blue!8}37.0                          & \cellcolor{blue!8}21.2                           & \cellcolor{blue!8}28.4                               & \cellcolor{blue!8}46.9                          & \cellcolor{blue!8}28.2                           & \cellcolor{blue!8}24.2                               & \cellcolor{blue!8}40.1                          & \cellcolor{blue!8}24.6                           \\ \hline
                                  & \checkmark                                                               & 29.5                               & 47.1                          & 29.7                           & 21.3                               & 39.9                          & 22.4                           & 31.5                               & 50.4                          & 31.0                           & 26.1                               & 42.3                          & 25.0                           \\
\multirow{-2}{*}{\begin{tabular}[c]{@{}c@{}}Fast-\\ RCNN\cite{girshick2015fast}\end{tabular}}       & $\times$                                                               & \cellcolor{blue!8}24.9                               & \cellcolor{blue!8}40.9                          & \cellcolor{blue!8}24.5                           & \cellcolor{blue!8}18.5                               & \cellcolor{blue!8}35.5                          & \cellcolor{blue!8}20.1                           & \cellcolor{blue!8}27.7                               & \cellcolor{blue!8}44.7                          & \cellcolor{blue!8}27.5                           & \cellcolor{blue!8}23.9                               & \cellcolor{blue!8}38.2                          & \cellcolor{blue!8}23.2                           \\ \hline
                                  & \checkmark                                                               & 30.6                               & 47.7                          & 30.1                           & 21.8                               & 40.5                          & 22.6                           & 29.1                               & 48.2                          & 28.8                           & 26.7                               & 42.6                          & 26.1                           \\
\multirow{-2}{*}{\begin{tabular}[c]{@{}c@{}}Deformable\\ DETR\cite{zhu2020deformable}\end{tabular}} & $\times$                                                               & \cellcolor{blue!8}26.1                               & \cellcolor{blue!8}41.8                          & \cellcolor{blue!8}25.9                           & \cellcolor{blue!8}20.8                               & \cellcolor{blue!8}36.2                          & \cellcolor{blue!8}20.3                           & \cellcolor{blue!8}28.3                               & \cellcolor{blue!8}45.6                          & \cellcolor{blue!8}27.9                           & \cellcolor{blue!8}24.4                               & \cellcolor{blue!8}38.7                          & \cellcolor{blue!8}23.1                           \\ \hline
                                  & \checkmark                                                               & 26.9                               & 46.5                          & 27.4                           & 22.7                               & 40.9                          & 23.0                           & 30.4                               & 48.6                          & 29.7                           & 25.9                               & 43.7                          & 26.3                           \\
\multirow{-2}{*}{\begin{tabular}[c]{@{}c@{}}Center-\\ Net\cite{law2018cornernet}\end{tabular}}       & $\times$                                                               & \cellcolor{blue!8}23.9                               & \cellcolor{blue!8}41.1                          & \cellcolor{blue!8}25.0                           & \cellcolor{blue!8}21.2                               & \cellcolor{blue!8}35.9                          & \cellcolor{blue!8}20.3                           & \cellcolor{blue!8}27.6                               & \cellcolor{blue!8}44.9                          & \cellcolor{blue!8}27.2                           & \cellcolor{blue!8}24.1                               & \cellcolor{blue!8}38.6                          & \cellcolor{blue!8}23.7                           \\ \hline
\end{tabular}
\end{table*}

\begin{figure*}
    \centering
    \includegraphics[scale=0.68]{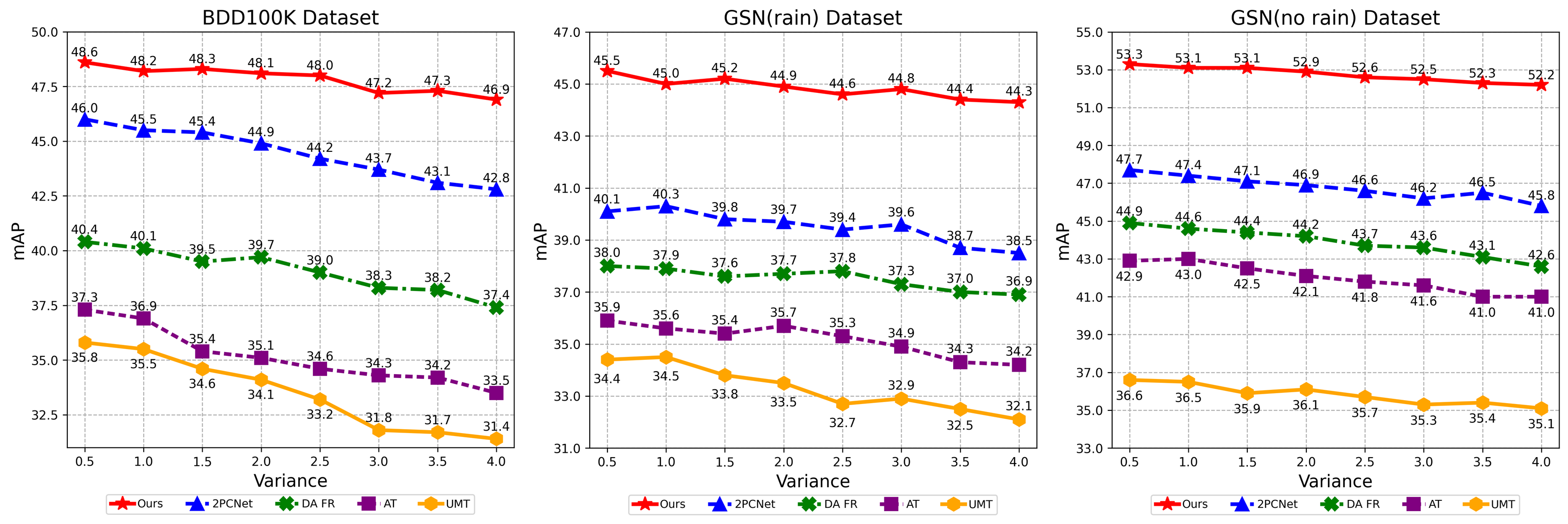}
    \caption{The results of different object detection methods with varying levels of Gaussian noise variance on BDD100K and GSN. Notably, the experimental results of the GSN dataset include both normal nighttime and rainy nighttime scenarios.}
    \label{fig:seg_rubost_all}
\end{figure*}

\textbf{GSN} Table \ref{GSN_results} reports the experimental results on the GSN dataset. To provide a more comprehensive and intuitive performance comparison, the results are divided into rain and no-rain conditions. In both scenarios, WTEFNet consistently achieves the best detection results. Specifically, under no-rain conditions, WTEFNet paired with YOLOv10 achieves the best performance with mAP@0.5 = 53.5, significantly outperforming other detection methods. It also attains the highest detection accuracy in multiple categories, including Pedestrian, Car, Bicycle, Traffic Light, and Traffic Sign. Compared to the second-best UIA method, WTEFNet improves mAP@0.5 by 3.0\%. Under rain conditions, WTEFNet with YOLOv10 achieves mAP@0.5 = 45.8, outperforming UIA by 2.5\%. Additionally, as observed in Table \ref{GSN_results}, low-light enhancement methods consistently provide more stable performance improvements in object detection under challenging low-light environments.

\subsection{Visualization Analysis}

\begin{figure}
    \centering
    \includegraphics[scale=0.35]{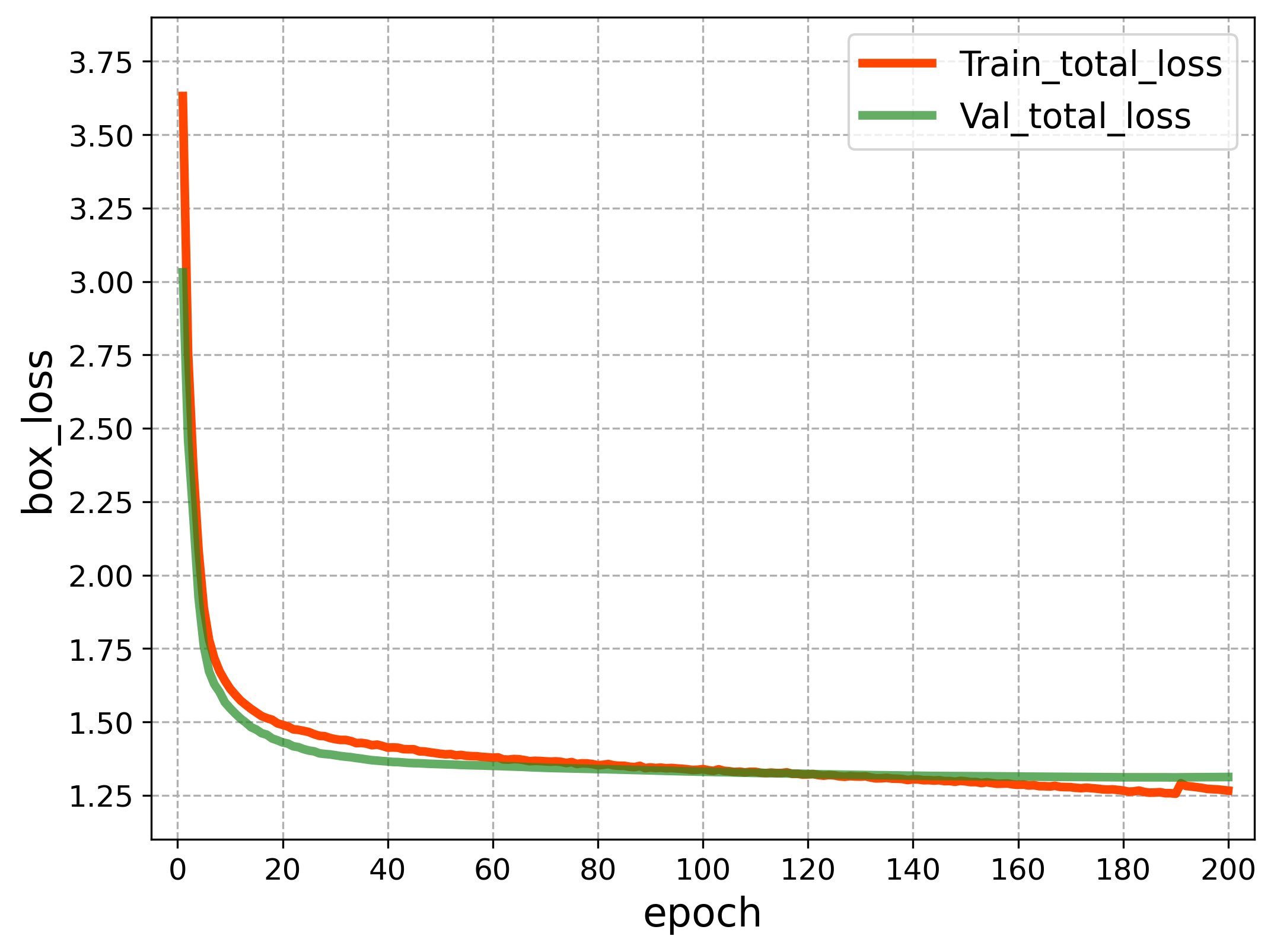}
    \caption{The loss function curves. The red and green curves indicate the progression of the loss function values on the training and validation sets, respectively, throughout the training process.}
    \label{fig:loss_epoch}
\end{figure}

\begin{figure}
    \centering
    \includegraphics[scale=0.35]{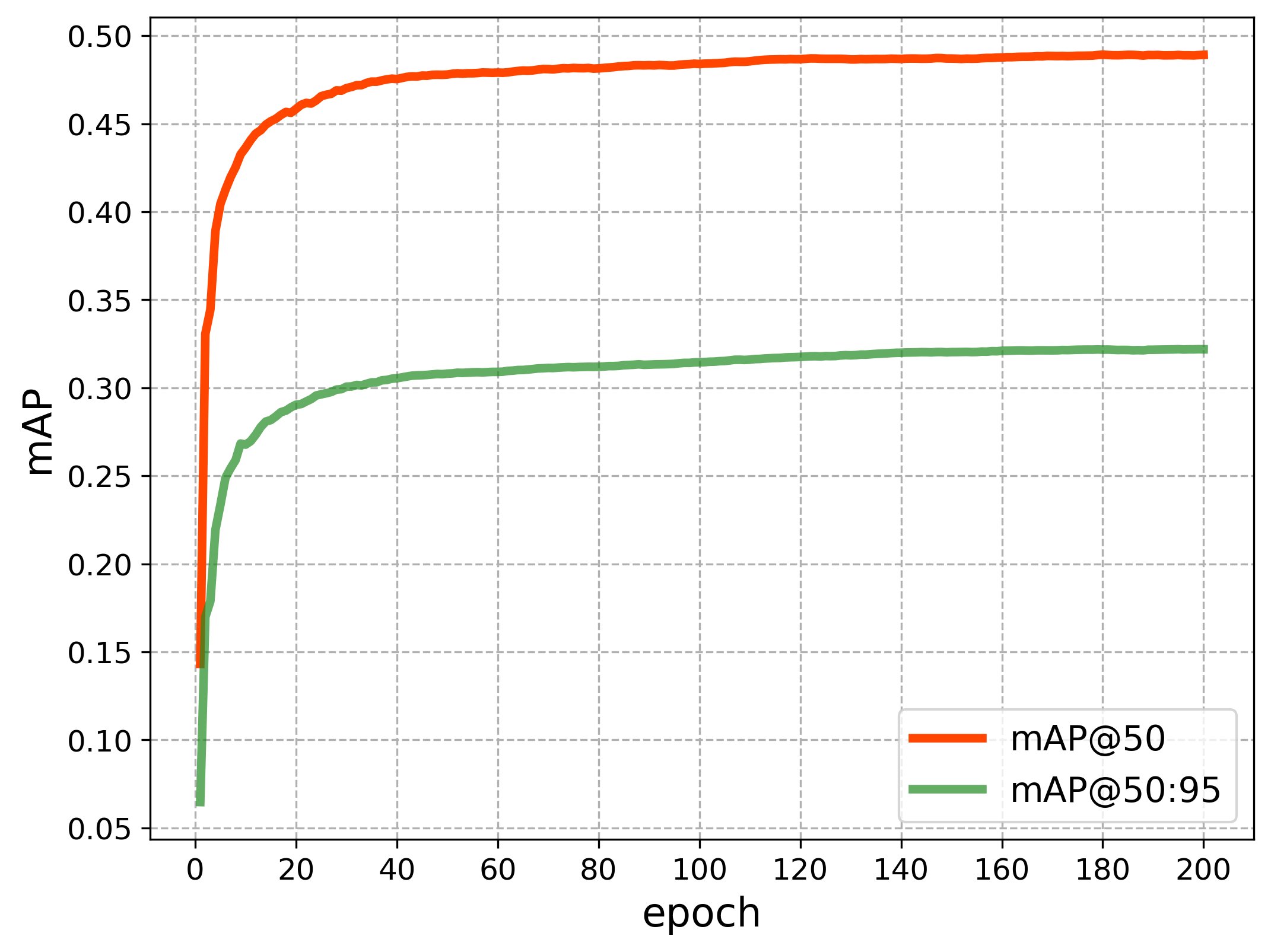}
    \caption{The mAP curves with different IoU thresholds. The red and green curves indicate the variation of mAP at the IoU threshold of 0.5 and the IoU range of [0.5:0.95], respectively. }
    \label{fig:map_epoch}
\end{figure}

To intuitively illustrate the superiority of WTEFNet more clearly, we visualize detection results from WTEFNet (paired with YOLOv10) and other comparison methods. Fig. \ref{fig:loss_epoch} and Fig. \ref{fig:map_epoch} illustrates several key performance metrics\cite{shirmohammadi2021machine} of WTEFNet during the training process. 

Fig. \ref{fig:visual_BDD100K} presents the detection results of various methods on the BDD100K dataset. We selected six representative images, labeled (a) to (f) from top to bottom. From left to right, the figure displays detection results from DA Fast-RCNN, UMT, WSA-YOLO, DAI-Net, 2PCNet, UIA and WTEFNet(Ours), along with the ground truth annotations. As observed, WTEFNet consistently achieves the best detection performance across different scenarios. For example, in scene (a), WTEFNet accurately detects challenging objects such as the Pedestrian (blue box) in the right region of the image and the Bus (red box) in the middle. In contrast, all other methods exhibit varying degrees of false positives or missed detections. In scene (c), numerous small, occluded objects appear in the central area, including Car (yellow box), Rider (dark green box), and Traffic Light (light green box). While WTEFNet exhibits a certain error rate, it still outperforms all other methods. In scene (f), only Retinexformer and WTEFNet correctly detect the black sedan in the lower right corner. Furthermore, WTEFNet uniquely provides accurate detections for the cluster of Cars (yellow box) in the left region and the Traffic Sign (purple box) in the center.

Fig. \ref{fig:visual_GSN} presents the visualized results of various methods on the GSN dataset. The first four images correspond to normal nighttime scenes (no-rain conditions), while the last two images are selected from rainy nighttime scenarios. In relatively well-lit conditions, such as scenes (c) and (d), most methods achieve reasonably accurate detection results, maintaining low false negative and false positive rates. However, in scenes (a) and (b), where illumination is severely insufficient, WTEFNet demonstrates significantly superior detection performance compared to other methods. For rainy nighttime scenes (e) and (f), additional challenges arise due to raindrops partially occluding objects and causing image blurring. Moreover, reflections on waterlogged roads introduce substantial interference, leading many methods to misidentify reflections as actual objects.

\subsection{Ablation Study}

To assess the generalization ability and adaptability of the proposed WTEFNet framework with mainstream object detection algorithms, as well as to explore how varying levels of wavelet transform influence detection performance, ablation studies were performed on the BDD100K and GSN datasets under the same experimental settings.

\textbf{Universal Test of WTEFNet} Table\ref{generalization_evaluation} presents the experimental results of several advanced object detection algorithms on the BDD100K, NuScenes, and GSN datasets. To clearly illustrate the improvements introduced by WTEFNet, for each method, we compare the performance with and without WTEFNet to highlight its contribution. As observed from the results, WTEFNet significantly enhances the performance of various detectors, including YOLOv10, YOLOv12, Fast R-CNN, Deformable DETR, and CenterNet, under low-light conditions. For the mAP@0.5 metric, some detection methods achieve performance gains of nearly 6\%. Even in rainy nighttime scenarios, WTEFNet provides a performance improvement of over 4\%. The experimental results confirm that WTEFNet possesses strong generalization ability, ensures compatibility with with various detectors, and effectively adapts to diverse low-light environments.

\begin{table}[h]
\centering
\caption{Ablation Study of Wavelet Transform Scales and Denoising Methods}
\label{ablation}
\setlength{\tabcolsep}{2.2pt} 
\renewcommand{\arraystretch}{1.2} 
\begin{tabular}{c|cccccc}
\hline
Model       & {\color[HTML]{333333} AP@0.5}        & {\color[HTML]{333333} AP@0.75}       & {\color[HTML]{333333} AP@0.5:0.95}   & APs                                  & APm                                  & APl                                  \\ \hline
w/o DWT      & 45.1                                 & 28.1                                 & 25.9                                 & 11.8                                 & 29.6                                 & 37.5                                 \\
DWT-scale-1 & 47.9                                 & 31.2                                 & 29.2                                 & 16.3                                 & 31.5                                 & 37.8                                 \\
DWT-scale-2 & {\color{blue} \textbf{48.9}} & {\color{blue} \textbf{31.4}} & {\color{blue} \textbf{31.8}} & {\color{blue} \textbf{17.5}} & {\color{blue} \textbf{32.3}} & {\color{blue} \textbf{38.8}} \\
DWT-scale-3 & 47.3                                 & 30.7                                 & 28.1                                 & 14.4                                 & 31.4                                 & 38.2                                 \\ \hline
\end{tabular}
\end{table}

\textbf{Wavelet-baesd Feature Extraction} We further investigated the impact of different wavelet transform scales on WTEFNet. The test results on the BDD100K dataset are shown in Table \ref{ablation}. Overall, the WFE module significantly improves detection performance, achieving the best results when the wavelet decomposition scale is set to 2.

\begin{figure}[]
    \centering
    \includegraphics[scale=0.45]{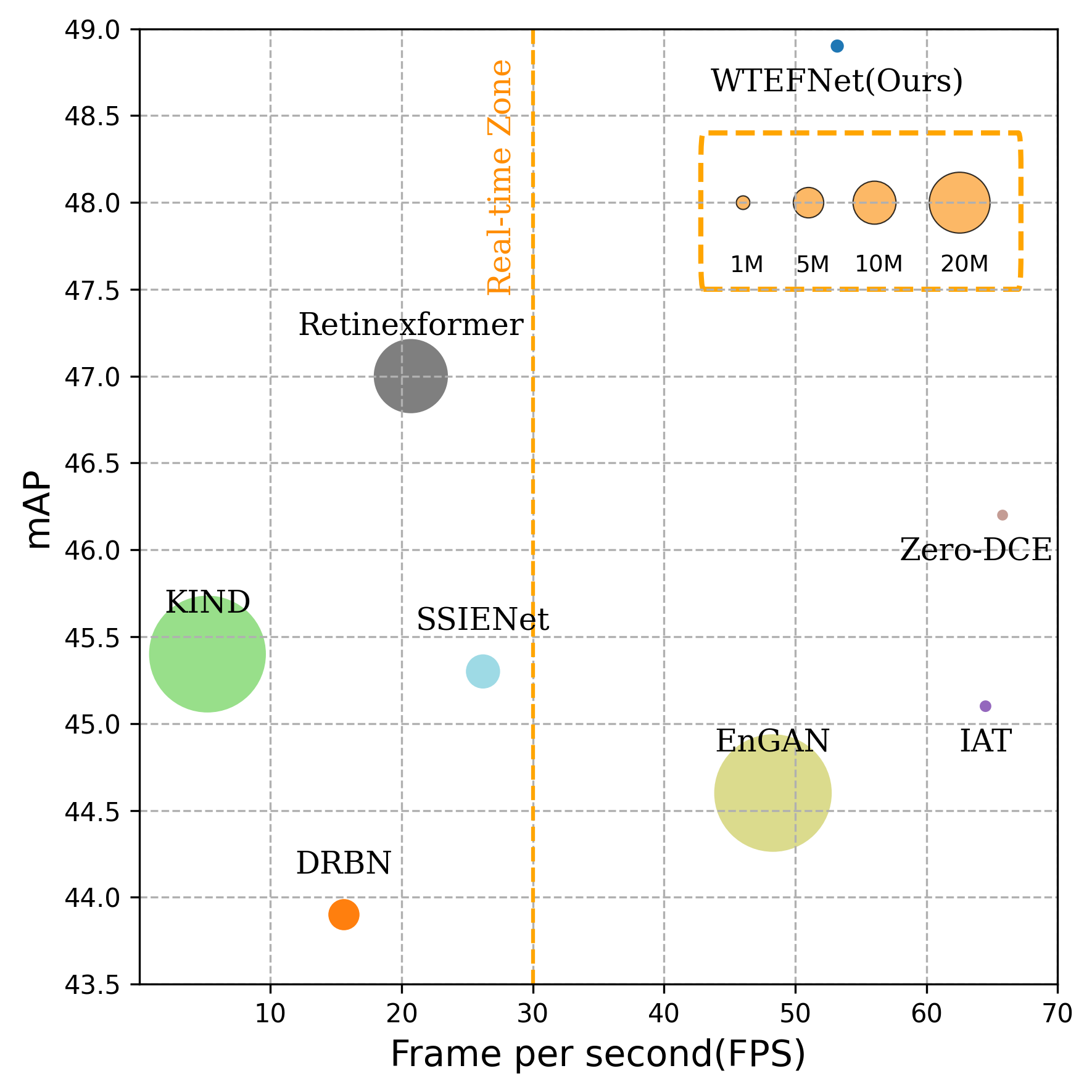}
    \caption{Performance comparison of different models. The size of the circle corresponds to the model's parameter quantity, with larger circles indicating a greater number of parameters.}
    \label{fig:map_fps}
\end{figure}

\subsection{Robustness Assessment}
The previous experiments have thoroughly verified the effectiveness of the proposed WTEFNet framework in object detection under diverse low-light environments. However, during vehicle operation, onboard cameras may be affected by noise interference. To further examine the model’s robustness in practical detection scenarios, we carried out noise sensitivity tests to evaluate its performance under different levels of noise disturbance. In particular, Gaussian noise with varying degrees of variance was added to the image data to mimic a range of noise conditions. As shown in Fig. \ref{fig:seg_rubost_all}, all methods experience varying degrees of performance degradation under Gaussian noise. However, compared to other approaches, the proposed WTEFNet exhibits the smallest performance drop on both datasets, demonstrating its superior robustness.

\subsection{Test on Embedded Devices}

\begin{figure}
    \centering
    \includegraphics[scale=0.5]{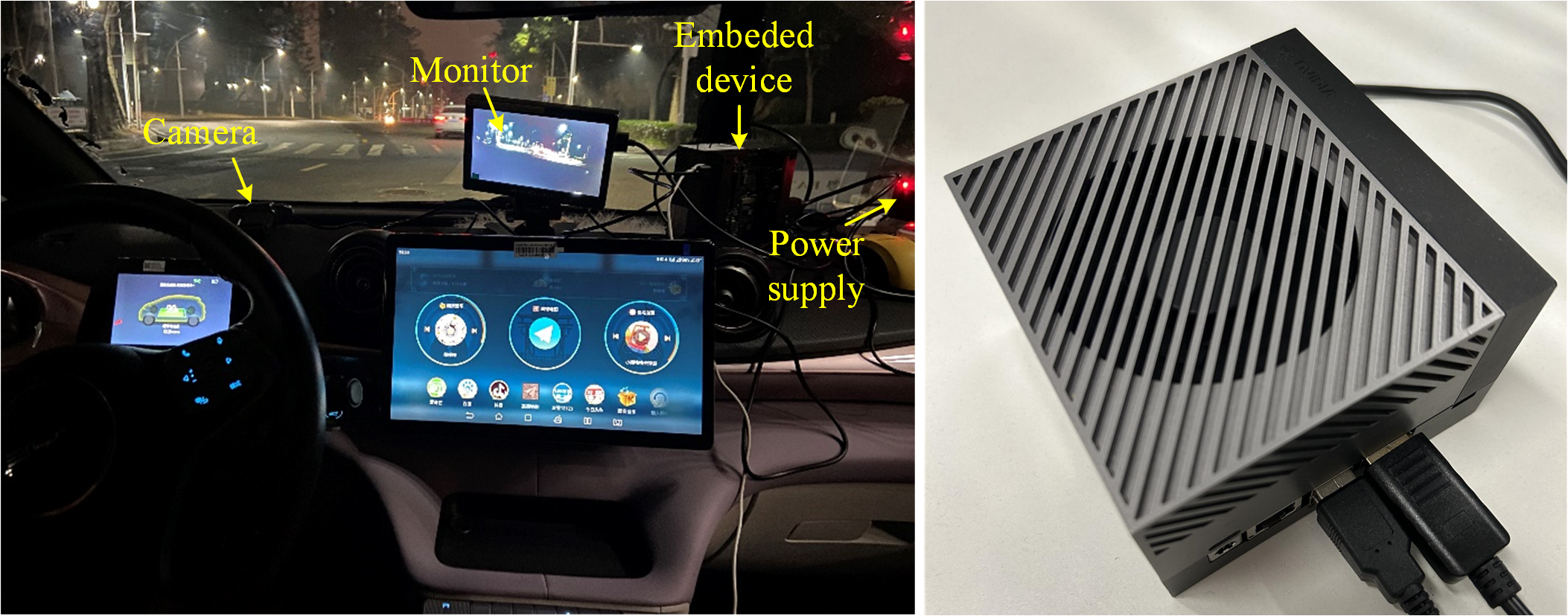}
    \caption{The energy-efficient embedded object detection instrumentation system equipped with NVIDIA Jetson AGX/Orin. }
    \label{fig:embebed_device}
\end{figure}

\begin{table}[h]
\centering
\caption{mAP and FPS of different methods on NVIDIA Jetson AGX/Orin.}
\label{device}
\setlength{\tabcolsep}{15pt} 
\renewcommand{\arraystretch}{1.2} 
\begin{tabular}{c|cc}
\hline
Methods       & mAP           & FPS           \\ \hline
EnGAN\cite{jiang2021enlightengan}         & 44.3          & 22.3          \\
SSIENet\cite{zhang2020self}       & 45.1          & 13.6          \\
Retinexformer\cite{cai2023retinexformer} & 46.4          & 11.7           \\
IAT\cite{cui2022you}           & 45.2          & \textbf{38.1} \\ 
WTEFNet(Ours)          & \textbf{48.1} & 30.4  \\ \hline
\end{tabular}
\end{table}

To further assess whether WTEFNet maintains a lightweight structure while satisfying the real-time perception demands of ADAS, we performed real-time performance evaluations of WTEFNet and several low-light image enhancement approaches on the NVIDIA Jetson AGX/Orin platform. Fig. \ref{fig:embebed_device} illustrates the real-world test scenarios and the embedded device used in the experiments, while Table \ref{device} presents the detailed results of the real-time performance evaluation. For a fair comparison, all methods adopt YOLOv10 as the detection head. The experimental results indicate that when deployed on the low-power embedded platform NVIDIA Jetson AGX/Orin, our model achieves the best detection accuracy while maintaining a high inference speed.

\section{Conclusion}

Conventional object detection methods are predominantly trained and evaluated on data captured under normal lighting conditions, making them vulnerable to missed or false detections in low-light environments. Such limitations can significantly compromise driving safety at night. Therefore, improving object detection accuracy and stability under low-light conditions is essential for ensuring safe nighttime driving, enabling reliable all-day perception, and advancing intelligent driving systems from driver assistance toward full autonomy.

In this work, we propose WTEFNet, a real-time object detection framework tailored for low-light conditions. It integrates low-light enhancement, wavelet-based feature extraction, and adaptive feature fusion to improve detection accuracy under challenging illumination. Furthermore, we introduced the GSN dataset, which provides high-quality annotations in diverse nighttime conditions, including rainy weather. Extensive experiments on multiple datasets confirm that WTEFNet achieves state-of-the-art performance and maintains real-time efficiency on embedded platforms, showing great potential for deployment in ADAS.

Future work will involve expanding low-light object detection datasets tailored for ADAS by collecting data samples from different cities under various weather conditions. Additionally, we aim to enhance the generalization capability of our detection algorithms.

\section*{Acknowledgements}
This project is jointly supported by National Natural Science Foundation of China (Nos.52172350, 51775565), Guangdong Basic and Applied Research Foundation (No.2022B1515120072), Guangzhou Science and Technology Plan Project (No.2024B01W0079), Nansha Key RD Program (No.2022ZD014), Science and Technology Planning Project of Guangdong Province (No.2023B1212060029). 

We gratefully acknowledge the support and assistance provided by Dr. Kening Li of the Shenzhen Institute of Information Technology.

\bibliographystyle{IEEEtran}

\bibliography{IEEEabrv,reference}
\begin{IEEEbiography}[{\includegraphics[width=1in,height=1.25in,clip,keepaspectratio]{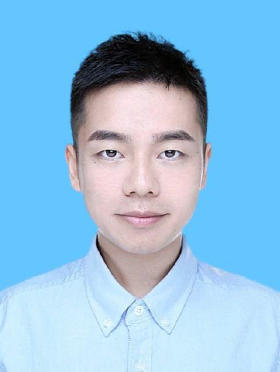}}]{Hao Wu} received the B.S. and M.S. degree in School of Civil Engineering and Tansportation from South China University of Technology in 2017 and 2020. He is currently pursuing the Ph.D. degree in Control Science and Engineering at Sun Yat-sen University, Shenzhen, 518107, Guangdong, China. His research interests include computer vision, wireless communications, and autonomous driving technologies.
\end{IEEEbiography}
\vspace{-1em}

\begin{IEEEbiography}[{\includegraphics[width=1in,height=1.25in,clip,keepaspectratio]{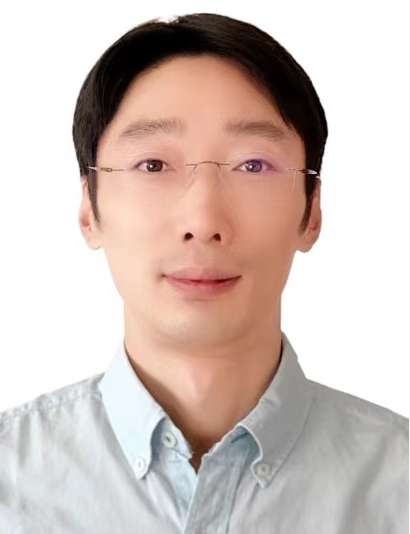}}]{Junzhou Chen} received his Ph.D. in Computer Science and Engineering from the Chinese University of Hong Kong in 2008, following his M.Eng degree in Software Engineering and B.S. in Computer Science and Applications from Sichuan University in 2005 and 2002, respectively. Between March 2009 and February 2019, he served as a Lecturer and later as an Associate Professor at the School of Information Science and Technology at Southwest Jiaotong University. He is currently an associate professor at the Guangdong Provincial Key Laboratory of Intelligent Transportation System, School of Intelligent Systems Engineering, Sun Yat-Sen University, Guangzhou 510275, China. His research interests include computer vision, machine learning, intelligent transportation systems, mobile computing and medical image processing.
\end{IEEEbiography}

\begin{IEEEbiography}[{\includegraphics[width=1in,height=1.25in,clip,keepaspectratio]{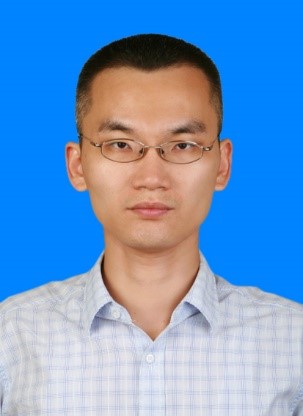}}]{Ronghui Zhang} received a B.Sc. (Eng.) from the Department of Automation Science and Electrical Engineering, Hebei University, Baoding, China, in 2003, an M.S. degree in Vehicle Application Engineering from Jilin University, Changchun, China, in 2006, and a Ph.D. (Eng.) in Mechanical $\And$ Electrical Engineering from Changchun Institute of Optics, Fine Mechanics and Physics, the Chinese Academy of Sciences, Changchun, China, in 2009. After finishing his post-doctoral research work at INRIA, Paris, France, in February 2011, he is currently an Associate Professor with the Guangdong Provincial Key Laboratory of Intelligent Transportation System, School of Intelligent Systems Engineering, Sun Yat-Sen University, Guangzhou 510275, China. His current research interests include computer vision, intelligent control and ITS.
\end{IEEEbiography}

\begin{IEEEbiography}[{\includegraphics[width=1in,height=1.25in,clip,keepaspectratio]{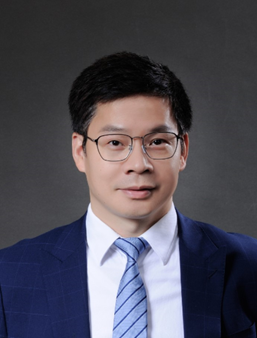}}]{Nengchao Lyu} is a professor of Intelligent Transportation Systems Research Center, Wuhan University of Technology, China. He visited the University of Wisconsin-Madison as a visiting scholar in 2008. His research interests include advanced driver assistance system (ADAS) and intelligent vehicle (IV), traffic safety operation management, and traffic safety evaluation. He has hosted 4 National Nature Science Funds related to driving behavior and traffic safety; he has finished several basic research projects sponsored by the National Science and Technology Support Plan, Ministry of Transportation, etc. He has practical experience in safety evaluation, hosted over 10 highway safety evaluation projects. During his research career, he published more than 80 papers. He has won 4 technical invention awards of Hubei Province, Chinese Intelligent Transportation Association and Chinese Artificial Intelligence Institute.
\end{IEEEbiography}
\vspace{-2em}

\begin{IEEEbiography}[{\includegraphics[width=1in,height=1.25in,clip,keepaspectratio]{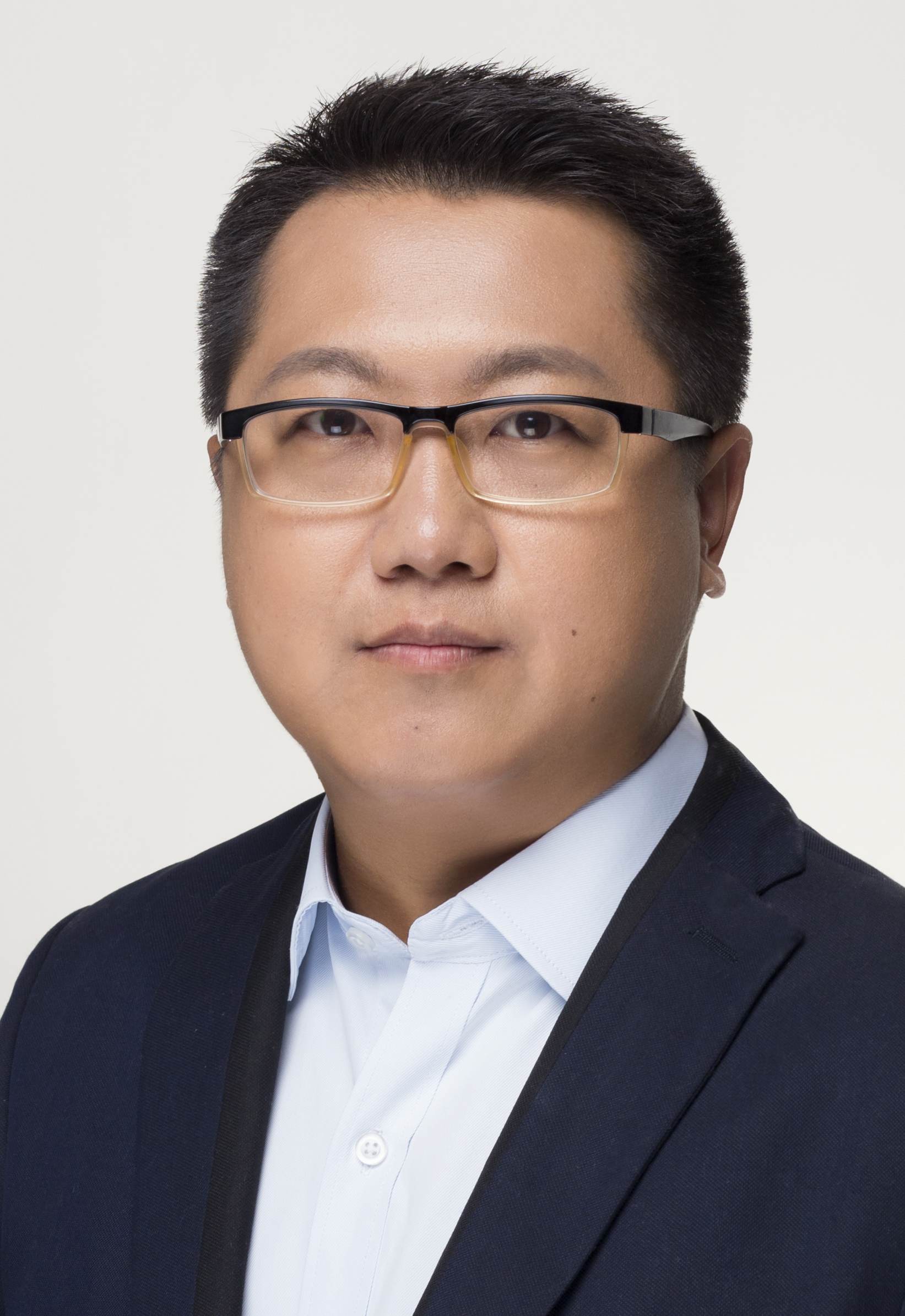}}]{Hongyu Hu} received the B.S. degree in computer science and technology in 2005, and the Ph.D. degree in traffic information engineering and control from Jilin University, Changchun, China in 2010. Since 2020, Dr. Hu has been a professor in State Key Laboratory of Automotive Simulation and Control. From 2019 to 2020, He was a visiting scholar at California PATH, UC Berkeley. Dr. Hu’s research interests include connected $\And$ automated vehicles, driver behavior analysis, advanced driver assistance system, and human-machine interaction. He is a committee member on Human Factors in Intelligent Transportation Systems and a member of Society of Automotive Engineers.
\end{IEEEbiography}
\vspace{-2em}

\begin{IEEEbiography}[{\includegraphics[width=1in,height=1.25in,clip,keepaspectratio]{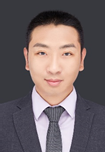}}]{Yanyong Guo} received the M.S. degree in transportation engineering from Chang’an University, Xi’an, China, in 2012, and the Ph.D. degree in transportation engineering from Southeast University, Nanjing, China, in 2016. From 2014 to 2015, he was a Visiting Ph.D. Student with The University of British Columbia. He is currently a Professor with the School of Transportation, Southeast University. His research interests include road safety evaluations, traffic conflicts techniques, advanced statistical techniques for safety evaluation, and pedestrian and cyclist behaviors. He received the China National Scholarship, in 2014, and the Best Doctoral Dissertation Award from the China Intelligent Transportation Systems Association, in 2017.\\[-1.5em]
\end{IEEEbiography}
\vspace{-2em}

\begin{IEEEbiography}[{\includegraphics[width=1in,height=1.25in,clip,keepaspectratio]{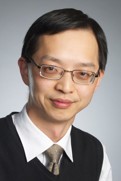}}]{ Tony Z. Qiu} is a Professor in the Faculty of Engineering at University of Alberta, Canada Research Chair Professor in Cooperative Transportation Systems, and Director of Centre for Smart Transportation. His research interest includes traffic operation and control, traffic flow theory, and traffic model analytics. He has published more than 180 papers in international journals and academic conferences, and has 7 awarded patents and 5 pending application patents. Dr. Qiu is working as the Managing Director for the ACTIVE-AURORA test bed network, which is Canadian National Connected Vehicle Test Bed, and sponsored by Transport Canada, City of Edmonton, Alberta Transportation and other funding agencies, to identify leading-edge Connected Vehicle solutions through research and development. His theoretical research has informed and enriched his many practical contributions, which have been widely supported by private and public sectors. Dr. Tony Qiu received his PhD degree from University of Wisconsin-Madison, and worked as a Post-Doctoral Researcher in the California PATH Program at the University of California, Berkeley before joining University of Alberta. Dr. Tony Qiu has been awarded Minister’s Award of Excellence in 2013, Faculty of Engineering Annual Research Award in 2015-2016, and ITS Canada Annual Innovation and R$\And$D Award in 2016 and 2017.

\end{IEEEbiography}

\end{document}